# CASR-Net: An Image Processing-focused Deep Learning-based Coronary Artery Segmentation and Refinement Network for X-ray Coronary Angiogram

Alvee Hassan[1,ǂ], Rusab Sarmun[2,ǂ], Muhammad E. H. Chowdhury[3,*], M Murugappan[4*,5], Abdulrahman Alqahtani[6,7], Balamurugan Balusamy[8*], Sohaib Bassam Zoghoul[9]

[1]Department of Biomedical Engineering, Military Institute of Science and Technology, Mirpur Cantonment, Dhaka 1216, Bangladesh.
[2]Department of Electrical and Electronic Engineering, University of Dhaka, Dhaka-1000, Bangladesh.
[3]Department of Electrical Engineering, Qatar University, Doha 2713, Qatar.
[4]Intelligent Signal Processing (ISP) Research Lab, Department of Electronics and Communication Engineering, Kuwait College of Science and Technology, Block 4, Doha, 13133, Kuwait.
[5]Department of Electronics and Communication Engineering, Vels Institute of Sciences, Technology, and Advanced Studies, Chennai, Tamilnadu, India.
[6]Department of Medical Equipment Technology, College of Applied, Medical Science, Majmaah University, Majmaah City 11952, Saudi Arabia
[7]Department of Biomedical Technology, College of Applied Medical Sciences in Al-Kharj, Prince Sattam Bin Abdulaziz University, Al-Kharj 11942, Saudi Arabia.
[8]School of Engineering and IT, Manipal Academy of Higher Education, Dubai Campus, Dubai, UAE
[9]Department of Radiology, Hamad Medical Corporation, Doha, Qatar.

ǂEqual contributors
*Corresponding author(s): M Murugappan (m.murugappan@kcst.edu.kw), Balamurugan Balusamy (balamurugan.balusamy@dxb.manipal.edu), Muhammad E. H. Chowdhury (mchowdhury@qu.edu.qa),

**Abstract**

Early detection of coronary artery disease (CAD) is critical for reducing mortality and improving patient treatment planning. While angiographic image analysis from X-rays is a common and cost-effective method for identifying cardiac abnormalities, including stenotic coronary arteries, poor image quality can significantly impede clinical diagnosis. We present the Coronary Artery Segmentation and Refinement Network (CASR-Net), a three-stage pipeline comprising image preprocessing, segmentation, and refinement. A novel multichannel preprocessing strategy combining CLAHE and an improved Ben Graham method provides incremental gains, increasing Dice Score Coefficient (DSC) by 0.31–0.89% and Intersection over Union (IoU) by 0.40–1.16% compared with using the techniques individually. The core innovation is a segmentation network built on a UNet with a DenseNet121 encoder and a Self-organized Operational Neural Network (Self-ONN) based decoder, which preserves the continuity of narrow and stenotic vessel branches. A final contour refinement module further suppresses false positives. Evaluated with 5-fold cross-validation on a combination of two public datasets that contain both healthy and stenotic arteries, CASR-Net outperformed several state-of-the-art models, achieving an IoU of 61.43%, a DSC of 76.10% and clDice of 79.36%. These results highlight a robust approach to automated coronary artery segmentation, offering a valuable tool to support clinicians in diagnosis and treatment planning.

## 1. Introduction

Coronary arteries (CAs) are vital vessels that encircle the heart and supply blood to the cardiac muscles. The two major coronary arteries are the left main coronary artery (LMCA) and the right coronary artery (RCA), originating from the aorta and dividing into smaller branches to supply blood within the heart. As coronary arteries directly provide oxygen-rich blood to the heart, any dysfunction or disease in these arteries can have catastrophic consequences, limiting the delivery of oxygen and nutrients to the heart muscle. The narrowing or blockage of coronary arteries, often caused by plaque formation, results in the development of stenotic arteries, known as coronary artery disease (CAD). CAD is responsible for 17.8 million deaths annually and remains one of the primary causes of mortality worldwide (Brown et al., 2022). Coronary artery disease can lead to chest pain (angina), shallow breathing patterns, irregular heart rhythms (arrhythmias), and even heart attacks (Clinic, 2022).

Cardiovascular diseases continue to be the leading cause of health complications globally (Roth et al., 2020), with coronary artery disease being the most prevalent cardiovascular disease (CVD) (Chowdhury et al., 2018). Early identification is crucial in preventing mortality from cardiovascular diseases such as heart attacks, which remain the principal cause of fatalities globally (Araújo et al., 2021).

X-ray coronary angiography (XCA) is the clinical gold standard for assessing coronary artery abnormalities. Automated segmentation of coronary vessels in XCA images offers valuable support for CAD diagnosis but faces persistent challenges. Image quality is often compromised by uneven illumination, low resolution, poor contrast, and a low signal-to-noise ratio, which hinder the detection of fine vascular details (Cervantes-Sanchez et al., 2019). Additional artifacts such as catheters, the diaphragm, and the spine can obscure vessel structures, while overlapping and tortuous arteries complicate boundary identification, especially in 2D projections (Gao et al., 2022). Although prior studies have explored a range of segmentation strategies (Cruz-Aceves et al., 2016; Gharleghi et al., 2022; Tian et al., 2021), accurate delineation of small and stenotic branches remains particularly difficult in conventional angiograms. Recent advances in computer vision have improved segmentation performance, but most approaches focus primarily on large vessels and contrast differences (Xian et al., 2020; Yang, Kweon, & Kim, 2019; Yang, Kweon, Roh, et al., 2019) often neglecting clinically significant diseased segments (Silva et al., 2021). Addressing this limitation is essential to achieve precise segmentation across both healthy and stenotic arteries for reliable clinical outcomes.Our study is motivated by persistent challenges in X-ray angiogram segmentation, including uneven illumination, poor artery-to-background contrast, low signal-to-noise ratio, and overlapping or twisted vasculature. These factors particularly hinder the accurate segmentation of small and stenotic branches, which are critical for clinical decision-making. To address these issues, we developed the Coronary Artery Segmentation and Refinement Network (CASR-Net), a pipeline based on an encoder–decoder structure.

Stenosis disrupts blood flow and often appears as discontinuities in angiograms. CASR-Net reconstructs these regions to restore vessel continuity while refining segmentation masks by eliminating misclassified pixels. A central innovation is the replacement of conventional Conv2D layers and ReLU activations in the UNet (Ronneberger et al., 2015) decoder with Self-ONN blocks and Tanh activations, which substantially improve continuity in narrow vessels. We further incorporated a multichannel preprocessing strategy that combines Contrast Limited Adaptive Histogram Equalization (CLAHE) (Zuiderveld, 1994) with a modified Ben Graham (Graham, 2015) method to enhance vessel visibility, suppress noise, and improve artery-to-background differentiation. As a final step, we implement a post-processing technique that reduces over-segmentation and under-segmentation by removing false positives through contour refinement stages and restoring vessel fragments via patch line generation techniques. By combining the two frameworks, accurate masks are created while fine vessel details are preserved, and background noise is filtered.

In summary, our unique segmentation pipeline, CASR-Net, demonstrates significant improvements in coronary artery study by thoroughly examining vessel characteristics and reducing segmentation errors. This innovative approach has the potential to advance the investigation and understanding of coronary artery pathologies with enhanced accuracy and reliability. In its entirety, the key contributions of this work are summarized as follows:

- The Coronary Artery Segmentation and Refinement Network (CASR-Net) introduce a novel encoder–decoder architecture that reduces breakage of slender vessel branches and enhances overall segmentation performance.
- A technical innovation replaces the traditional Conv2d layer and ReLU activation with a Self-ONN block and Tanh activation, minimizing continuity loss in narrow vessels and improving segmentation accuracy.

- An incremental improvement is achieved through a multichannel image enhancement technique combining CLAHE and Ben Graham features, which enhances artery-to-background differentiation and vessel clarity in X-ray angiograms.
- A set of incremental post-processing refinements further improves prediction quality by reducing false positives and false negatives, leading to more robust segmentation.

The remainder of this paper is organized as follows. Section 2 reviews related literature on preprocessing, segmentation, and post-processing methods for coronary artery analysis. Section 3 outlines the motivation for this study. Section 4 details the experimental setup, including datasets, preprocessing strategies, and training protocols. Section 5 presents the proposed methodology, followed by Section 6, which reports the results and discusses the performance of CASR-Net in comparison with existing approaches. Section 7 highlights the limitations and potential future directions, and Section 8 concludes with key findings and clinical implications.

## 2. Literature Review

### 2.1 Preprocessing Techniques

Preprocessing plays a vital role in coronary artery segmentation because the quality of XCA images directly affects vessel delineation. Common challenges include low contrast, uneven illumination, and noise that can obscure fine vascular details. CLAHE is widely used to improve local contrast by applying histogram equalization on non-overlapping tiles. Prior work has shown that CLAHE enhances visibility of narrow arteries while limiting noise amplification, which is particularly useful for angiograms with poor contrast(Iyer et al., 2021; Jiang et al., 2021) Another technique adapted to coronary angiograms is Ben Graham's preprocessing method, originally developed for retinal images. By subtracting the local average color, it accentuates vascular structures against the background. In coronary angiograms, direct application can yield circular patch artifacts, so in this study the method is modified to mitigate patch related shadowing and improve uniformity in regions with complex vasculature (Graham, 2015).

Several other preprocessing techniques have proven effective for coronary artery enhancement beyond CLAHE and Ben Graham's method. Several morphological operations, particularly the top-hat transform, have been used for vessel enhancement by removing small elements and details while suppressing background structures (Eiho & Qian, 1997; Tao et al., 2022). Vesselness filters form another cornerstone in vascular imaging. Frangi's vesselness filter, based on Hessian matrix eigenvalue analysis, has become a cornerstone technique for tubular structure enhancement, identifying vessel-like patterns while suppressing non-vascular elements (Li et al., 2024). Extensions such as Sato's filter and multi-scale Hessian-based variants improve sensitivity to vessels of varying sizes, yet they can amplify noise in challenging angiographic data. Recursive filtering enhances edges and reduces noise, but tends to miss very thin branches, while gradient vector flow (GVF) improves boundary detection at the cost of high computational expense. Gabor filters, designed to capture oriented patterns at multiple scales, have also been applied for vessel detection, with parameter optimization yielding notable improvements (Cruz-Aceves et al., 2016) However, their computational burden and limited robustness in stenotic regions restrict broader adoption.

Beyond CLAHE and Ben Graham's method, several preprocessing techniques have been widely explored for coronary artery enhancement. Morphological operations, particularly the top-hat transform, have been used to highlight small bright structures such as vessels on dark backgrounds, though they remain sensitive to noise structures. The Frangi filter, based on Hessian matrix eigenvalue analysis, effectively detects elongated tubular patterns while suppressing non-vascular structures, but often struggles with low-contrast (Li et al., 2024).While these classical approaches provide valuable enhancement strategies, their limitations in handling low contrast, high noise, and stenotic regions of coronary angiograms reduce their reliability in clinical applications. In this work, we adopt a multichannel preprocessing pipeline that integrates CLAHE with an enhanced Ben Graham method, specifically designed to mitigate uneven illumination, suppress background noise, and improve both local and global contrast. This combination provides superior vessel visibility compared to conventional vesselness filters, making it more effective in addressing the challenges of X-ray coronary angiography.

**2.2 Segmentation Methods**

A wide range of methods have been proposed for coronary artery segmentation in X-ray angiograms (XCA), including model-based approaches, pattern-recognition techniques (Tayebi et al., 2013) tracking algorithms (Lesage et al., 2008; Zhou et al., 2008) classical machine learning methods (Cervantes-Sanchez et al., 2019; Cervantes-Sanchez et al., 2020; Friedman, 2001; Gao et al., 2022; *Medical Image Computing and Computer-Assisted Intervention – MICCAI 2015*, 2015; Shi et al., 2020; Shuvo et al., 2021) and more recently, deep learning strategies (Chen et al., 2014; He et al., 2016; Huang et al., 2017; Szegedy et al., 2017; Yang, Kweon, & Kim, 2019; Yang, Kweon, Roh, et al., 2019; Zhou & Feng, 2017, 2019).

Deep learning methods, particularly convolutional neural networks (CNNs), have shown the greatest promise, treating vessel segmentation as a pixel-wise classification problem (Nasr-Esfahani et al., 2018). Lightweight networks such as MobileNet (Howard et al., 2019; Howard et al., 2017; Sandler et al., 2018) bottleneck residual architectures, and compact designs like CNL-UNet (Shuvo et al., 2021) have been proposed to balance accuracy with computational efficiency. More advanced strategies incorporate attention mechanisms to capture long-range spatial dependencies and refine vessel boundaries. Recent innovations include the Progressive Perception Learning (PPL) framework, which integrates context, interference, and boundary perception modules to enhance segmentation across multiple image granularities, achieving Dice scores above 95% (Hongwei Zhang et al., 2022). Similarly, the Context Interactive Deep Network (CIDN) strengthens vessel edge delineation by linking low- and high-level features. CIDN introduces a Bio-inspired Attention Block (BAB), which leverages wavelet transforms for orientation-aware learning, and a Multi-scale Interactive Block (MIB) for optimized spatial representation. It further employs a compound loss that combines binary cross-entropy with active contour elasticity to better handle complex vessel boundaries (Zhang et al., 2024).

Generative Adversarial Networks (GANs) have also been applied to coronary segmentation. For example, UENet uses a U-Net generator with a pyramid-structured discriminator to improve vessel connectivity (Shi et al., 2020), while ASCARIS incorporates multi-constraint preprocessing and an attention-based nested U-Net to achieve superior performance (Algarni et al., 2022a). Other approaches, such as PSPNet (Zhu et al., 2021) and ensemble frameworks combining Deep Forest classifiers with gradient boosting (Gao et al., 2022), have also reported competitive results. Parallel to CNN-based architectures, transformer-based models have gained traction in medical image segmentation. TransUNet combines the global context modeling of transformers with the precise localization capabilities of U-Net, producing strong results across cardiac imaging tasks (Chen et al., 2021). Variants such as large-kernel attention modules (Liu et al., 2023), CiT-Net (Lei et al., 2023), and Tim-Net (Hongbin Zhang et al., 2022) which integrates convolutional priors into transformer blocks, further enhance performance on small-scale medical datasets. Recently, MedSAM, adapted from the Segment Anything Model (SAM), demonstrated competitive performance across multimodal segmentation tasks, highlighting the growing influence of foundation models in medical imaging (Ma et al., 2024).

The current state of the art in medical segmentation is increasingly defined by hybrid architectures that integrate GANs and transformers. For example, CASTFormer combines adversarial training with transformer-based modules to improve robustness in 2D medical imaging (You et al., 2022). In coronary segmentation, DR-LCT-UNet leverages Dense Residual modules with Local Contextual Transformer blocks, achieving a Dice score of 85.8% on CCTA data (Wang et al., 2023). These hybrids represent a promising direction, combining the structural learning capacity of transformers with the connectivity-preserving advantages of GANs.

Finally, alternative strategies such as patch-based training have also been effective. (Cervantes-Sanchez et al., 2020) showed that patch-level learning enhances U-Net performance on narrow branches, while (Dong et al., 2023) introduced CAS-Net, a 3D network with attention-guided feature fusion (AGFF) and scale-aware enhancement (SAFE) modules for better semantic capture. Together, these developments underscore the shift from classical methods toward advanced, hybrid deep learning architectures that increasingly integrate attention, adversarial learning, and transformer modules to achieve robust coronary artery segmentation.

### 2.3 Post-Processing Techniques

Coronary artery segmentation in X-ray angiograms remains challenging due to the complex and narrow geometry of the arteries, motion artifacts from cardiac activity, calcification-induced stenosis, and overall poor image quality (Meng et al., 2009). Even with advanced deep learning models, post-processing is often required to refine outputs and ensure reliable clinical applicability. One of the most widely adopted strategies is contour refinement, which aims to eliminate false positives. Small contours are identified based on a size threshold, inverted, and removed using bitwise operations with the original mask, resulting in a cleaner and more accurate representation of the vasculature (Mulay et al., 2021). Another important technique is patch line generation, designed to restore continuity in fragmented thin branches. This method applies morphological thinning to create skeletonized masks, detects endpoints using convolution, and connects them with patch lines guided by Euclidean distance, significantly reducing vessel discontinuities.

Beyond these structural corrections, other post-processing approaches target common sources of segmentation error. Noise suppression and contrast adjustment are frequently used to remove residual artifacts and improve edge definition. Vesselness-enhancing diffusion filters have been applied to strengthen the continuity of tubular structures (Gao et al., 2022), while Hessian-based edge detection combined with multi-scale vessel region detection has proven particularly effective in stenotic cases (Fazlali et al., 2015). In addition, strategies such as topology-aware loss designs and focal variants of the Dice or Tversky index (Abraham & Khan, 2019) can indirectly improve post-segmentation refinement by penalizing missed thin branches or false positives during training. More recently, synthetic data augmentation with GANs has emerged as a complementary post-processing aid. By generating realistic but diverse training examples, GAN-based augmentation improves the robustness of CNNs to variability in vessel appearance. For example, (Bernard et al., 2018) demonstrated higher sensitivity and specificity in lesion classification with GAN-augmented datasets, a result with clear implications for coronary angiography.

Taken together, these post-processing techniques highlight the importance of refining segmentation outputs beyond the initial prediction stage. By removing spurious regions, reconnecting disjoint branches, and enhancing vessel continuity, such methods play a crucial role in producing clinically reliable coronary artery segmentations.

Table 1 summarizes a comparative analysis obtained from the literature that involves the segmentation of coronary arteries from XCA images associated with the Database X-ray Coronary Angiograms (DCA1). The matrices selected for the comparison are Accuracy (Acc), Intersection over Union (IoU), Dice Score/Dice Similarity Co-efficient (DSc/DSC), Sensitivity (SN), and Specificity (SP), in line with our study.

**Table 1**. A brief performance comparison of state-of-the-art segmentation methods for the DCA1 X-ray coronary angiographic dataset

| S.No | Year | Method | Dataset | Fold Strategy | Acc | IoU | DSC | SN | SP |
|---|---|---|---|---|---|---|---|---|---|
| 1. | 2021 | U-net with a novel Loss Function (Araújo et al., 2021) | DCA1 | - | 0.97 | - | - | - | - |
| 2. | 2019 | Multilayer Perception (MLP) Architecture (Cervantes-Sanchez et al., 2019) | DCA1 | Training Set: 100 images, Test Set: 30 images | 0.9698 | - | 0.6857 | 0.6364 | 0.988 |
| 3. | 2022 | Bottleneck Residual U-Net (BRU-Net)(Tao et al., 2022) | DCA1 | 85% for training data, 10% for validation data and 5% for test data | 0.9789 | - | - | 0.877 | 0.9789 |
| 4. | 2021 | Vessel Specific Skip chain Convolutional Network (VSSC Net) (Samuel & Veeramalai, 2021) | DCA1 | - | 0.97 | - | - | 0.7728 | 0.9809 |
| 5. | 2020 | Patch-based training of U-Net (Cervantes-Sanchez et al., 2020) | DCA1 | 100 were used for training, and the other 30 for testing | 0.977 | - | 0.779 | 0.988 | 0.773 |
| 6. | 2021 | Multiresolution and Multiscale Convolution Filtering based U-Net (Jiang et al., 2021) | DCA1 | 6-fold cross-validation | 0.9765 | - | 0.7905 | 0.7978 | 0.9885 |
| 7. | 2021 | Edge Adaptive Instance Normalization (Edge-AdaIN) Style Transfer Network (Mulay et al., 2021) | DCA1 | 100 were used for training, and the other 30 for testing | 0.9658 | - | 0.7165 | 0.7867 | 0.9756 |
| 8. | 2022 | Diffusion adversarial representation learning (DARL) Network (Kim et al., 2022) | DCA1 | - | - | 0.427 | 0.572 | - | - |
| 9. | 2022 | Full Resolution U-Net (FR-UNet) (Liu, Yang, Tian, Cao, et al., 2022) | DRIVE, CHASE_DB1, STARE, DCA1 | 85% for training, 10% for validation, and 5% for testing | 0.9788 | 0.6708 | - | 0.8248 | 0.9875 |
| 10. | 2022 | Residual-Attention UNet++ (Li et al., 2022) | DCA1 | 100 were used for training, and the other 30 for testing | - | 0.6657 | 0.7248 | 0.8335 | - |
| 11. | 2024 | Context Interactive Deep Network (CIDN)(Zhang et al., 2024) | DCA1 | - | 0.9795 | - | - | 0.8919 | 0.9830 |
| 12. | 2022 | Multiscale Attention Aggregation Network (MAA-Net) (Liu, Yang, Tian, Pan, et al., 2022) | DCA1 | training set of 100 images and a test set of 34 | 0.9785 | - | - | 0.8545 | 0.9854 |
| 13. | 2025 | Multi-attention dynamic sampling network (Multi-ADS-Net): Cross-dataset pre-trained model for generalizable vessel segmentation in X-ray coronary angiography (Wu et al., 2025) | ARCADE, XCAD, SVS, DCA1, CHUAC | 5-fold cross-validation | - | - | 0.819 | - | - |
| 14. | 2024 | Optimizing ensemble U-Net architectures for robust coronary vessel segmentation in angiographic images (Chang et al., 2024) | DCA1 | 5-fold cross validation | 0.97 | - | 0.76 | - | - |
| 15. | 2024 | CAS-GAN: A Novel Generative Adversarial Network-Based Architecture for Coronary Artery Segmentation (Hamdi et al., 2024) | CORONAR, DCA1, CHUAC | training set of 100 images and a test set of 34 | - | - | 0.8245 | - | - |

With respect to existing studies, our proposed method introduces key advancements that address the persistent challenges in coronary artery segmentation from X-ray angiograms. Previous work has attempted to mitigate geometric difficulties in thin vessels; for example, (Cervantes-Sanchez et al., 2020) employed a patch-based training strategy to improve continuity in narrow branches. While such approaches demonstrate the value of localized learning, they often fail to generalize to poor-quality images or stenotic regions. In contrast, our method adopts a fundamentally different strategy by integrating a Self-ONN-based decoder within a U-Net framework. Although Self-ONNs have been applied successfully in classification tasks in the literature (Hossain et al., 2023; Kiranyaz et al., 2015; Qin et al., 2024; Rahman et al., 2024), their use in segmentation pipelines remains limited. Here, the Self-ONN decoder allows the network to better adapt to the complex and non-linear geometry of coronary arteries, thereby improving vessel continuity and reducing breakages in narrow branches. In addition, we introduce a multichannel preprocessing technique that combines CLAHE with a modified Ben Graham method to address uneven illumination and poor contrast. Unlike prior studies that primarily rely on single-channel enhancement, our multichannel approach substantially improves artery-to-background separation, particularly in challenging stenotic regions.

## 3. Motivation

Accurate diagnosis and treatment of cardiovascular disease rely heavily on the ability to identify and segment coronary arteries from medical images, particularly XCAs. Reliable segmentation is essential for clinicians to evaluate the severity and location of stenosis, which directly influences patient management and therapeutic decision-making. Although prior studies have introduced promising techniques for coronary artery segmentation, significant challenges remain. Chief among these is the accurate delineation of stenotic regions in poor-quality images and the preservation of continuity in narrow arterial branches. These areas remain underexplored despite their clear clinical importance. The risks associated with stenotic coronary arteries depend not only on the degree of narrowing but also on its anatomical location and the patient's overall condition, making precise detection critical for CAD assessment.

This study proposes a comprehensive pipeline designed to address these challenges. The approach is capable of processing low-quality XCA images, separating arteries from the background, and producing refined, clinically reliable vessel segmentations. Importantly, the use of a diverse dataset containing stenotic arteries enables the model to learn from these cases and improve performance in clinically challenging regions. A novel segmentation network is also introduced, specifically designed to minimize breakages in thin branches while maintaining accuracy across stenotic areas. By targeting both vessel continuity and robustness to poor imaging conditions, the proposed framework aims to enhance the diagnostic process and provide clinicians with more precise tools for evaluating CAD severity.

## 4. Experimental Setup

### 4.1 Dataset Description

A number of public datasets are available for coronary artery segmentation, each with distinct characteristics. The ARCADE dataset provides vessel-tree and plaque annotations from more than 1,200 X-ray angiograms (Popov et al., 2023) Other datasets contribute complementary perspectives, such as the coronary dominance classification dataset (Zreik et al., 2019), Mahmoudi et al.'s angiograms with SYNTAX scores (Mahmoudi et al., 2025) and CADICA invasive angiography videos enriched with clinical metadata (Jiménez-Partinen et al., 2024)For non-invasive imaging, ImageCAS offers a benchmark for coronary CT angiography segmentation (Zeng et al., 2023).

In this study, we specifically selected the DCA1 dataset (Cervantes-Sanchez et al., 2019) and the Angiographic Dataset for Stenosis Detection (Danilov et al., 2021). Because unlike the ARCADE dataset which provides disjoint branch masks, the DCA1 provides complete artery masks that serve as a strong baseline for learning normal vessel anatomy, while the Stenosis Detection dataset by Danilov et al. introduces clinically diverse stenotic cases acquired under heterogeneous imaging conditions. Together, these two datasets offer a complementary balance between structural completeness and pathological diversity, directly aligning with our objective of improving segmentation performance in challenging 2D XCA conditions.

### 4.1.1 X-ray Coronary Angiograms (DCA1)

The DCA1 dataset comprises 134 X-ray coronary angiograms, accompanied by corresponding ground-truth vessel segmentations meticulously annotated by a (Cervantes-Sanchez et al., 2019).Sourced from the Cardiology Department of the Mexico Social Security Institute, UMAE T1-León, each angiogram was originally a 300 × 300 pixel PGM image, subsequently processed into a 256 × 256 PNG format for this study. The imaging conditions are relatively consistent throughout the dataset, with minimal variability in acquisition parameters. A key characteristic of DCA1 is its predominant representation of normal coronary artery anatomy, with only a limited number of pathological segments. This relative lack of disease complexity makes it a valuable resource for training a model on baseline vessel structure, though its main limitation is the under-representation of severe disease.

### 4.1.2 Angiographic Dataset for Stenosis Detection

CLAHE (Danilov et al., 2021).. These data were acquired at the Research Institute for Complex Issues of Cardiovascular Diseases. A significant feature of this dataset is its heterogeneity; images were captured on different systems (Coroscop, Siemens; Innova, GE Healthcare) and vary in resolution, with sizes ranging from 512 × 512 to 1000 × 1000 pixels. All patients had functionally significant stenosis, defined as ≥70% diameter stenosis by Quantitative Coronary Analysis (QCA) or 50–69% stenosis with functional evidence of ischemia. An interventional cardiologist carefully selected frames that clearly showed contrast passage through the stenotic region. This resulted in a total of 8,325 grayscale images featuring visible stenotic segments with bounding box-based lesion annotations, reflecting a spectrum of clinical severity. The diversity of lesion sizes was notable, categorized as small (30%), medium (69%), and large (1%). The primary limitation of this dataset is its exclusive focus on single-vessel disease, which provides a controlled environment for stenosis detection but does not capture the complexity of multi-vessel disease.

Since the original dataset was designed for stenosis detection rather than full vessel segmentation, we adopted a two-step approach to obtain consistent ground-truth masks. First, the segmentation model trained on the DCA1 dataset was used to generate initial masks for one frame of each video in the Danilov dataset. These preliminary outputs were then manually refined under the close supervision of three experienced radiologists to ensure accuracy, consistency, and reliability across cases. This procedure allowed us to leverage the stenosis-rich diversity of the Danilov dataset while maintaining annotation quality comparable to DCA1, thereby creating a robust dataset suitable for evaluating CASR-Net in clinically challenging stenosis scenarios. In total 214 images were annotated and utilized from this dataset.

### 4.1.3 Combined Dataset

The two XCA datasets were combined in this study to include both healthy and stenotic coronary arteries, resulting in a combined dataset of 348 PNG images, each sized at 256 × 256 pixels, along with their corresponding masks. The dataset was split into training (80%) and testing (20%) sets, with the validation set (20%) formed from the training set. To evaluate the effectiveness of the coronary artery segmentation technique, k-fold cross-validation (k=5) was performed. This method is widely used to assess the performance of classification systems and is recommended for reliable accuracy estimation with low variation (Wong & Yeh, 2020) Click or tap here to enter text.. Each test fold consists of approximately 70 images, with each angiography exclusively assigned to one-fold. This technique ensures the generalization of the study for independent angiography datasets.

### 4.2 Pre-processing

In this study, image processing techniques were employed to address issues related to non-uniform illumination and poor contrast ratios in angiograms. Creating a reliable segmentation algorithm becomes challenging due to variations in image acquisition techniques and the presence of diseases. Before inputting images into the model, image enhancement algorithms were used to enhance the visibility of arteries. Two primary image processing techniques, namely the CLAHE method and the Ben Graham image processing, were utilized to differentiate arteries from poorly contrasted backgrounds (Dey, 2020).

#### 4.2.1 CLAHE Method

Contrast Limited Adaptive Histogram Equalization (CLAHE) was applied to improve vessel visibility in angiograms affected by non-uniform illumination and poor contrast (Dey, 2020). CLAHE enhances local contrast by dividing the image into non-overlapping tiles and applying histogram equalization within each tile. To prevent over-amplification of noise, a clip limit is imposed on the histogram. Bins exceeding this threshold are clipped, and the excess counts are redistributed evenly among all bins (Iyer et al., 2021b; Jiang et al., 2021b; Samuel & Veeramalai, 2021b; Tao et al., 2022b). The redistribution is expressed as:

$$H' = min(H, C_{\text{limit}}) + \frac{\sum_{i=1}^{N} max(0, H_i - C_{\text{limit}})}{N} \quad (1)$$

Where $H$ is the original histogram, $H'$ is the modified histogram, N is the number of bins in the histogram, and $H_i$ represents the count of the $i^{th}$ bin. The process ensures that the contrast enhancement is limited, reducing the risk of amplifying noise in the image.

The choice of grid size and clip limit is critical. Small grids or low clip limits may suppress low-frequency illumination but also risk erasing thin or low-contrast vessels. Larger grids or high clip limits can preserve fine detail but may amplify noise or produce block artifacts at tile borders. To determine the optimal parameters, an ablation study was conducted using grid sizes of 4×4, 8×8 and 12×12, with clip limits ranging from 4 to 12. Both quantitative segmentation metrics and visual inspections of vessel clarity were evaluated. A grid size of 4×4 combined with a clip limit of 8 consistently achieved the best balance between vessel enhancement and noise suppression (see **Supplementary Table 1**)

Although CLAHE improves local contrast substantially, it is not without limitations. At tile borders, tile-based equalization may introduce blocky artifacts or halo effects, and reshaping local histograms may distort apparent vessel calibers. Moreover, results vary across imaging devices, exposure settings, and pathologies depending on parameter selection. The CLAHE preprocessing step, in spite of these limitations, enhances the visibility of coronary arteries in angiograms, particularly in low-contrast areas.

#### 4.2.2 Improved Ben Graham

To further enhance the visibility of vessels against the background, a novel and improved version of the well-established Ben Graham image enhancement technique was implemented. In the Kaggle Diabetic Retinopathy competition in 2015, Ben Graham utilized an image enhancementmethod by subtracting the image's local average color, making vessels more prominent against the background (Graham, 2015).. This technique involves three key steps aimed at reducing intra-class variation caused by extraneous factors such as lighting and camera quality. Firstly, each image undergoes a local average color subtraction where the mean color value of the image is calculated and then subtracted from each pixel, effectively centering the color distribution. This operation can be represented by the equation:

$$P'_{i,j} = P_{i,j} - \frac{1}{MN} \sum_{m=1}^{M} \sum_{n=1}^{N} P_{m,n} \tag{2}$$

Where $P'_{i,j}$ is the new pixel value at coordinates i, j, $P_{i,j}$ is the original pixel value and M, N are the dimensions of the image. The second step involves scaling the local average to 50% gray, to maintain a standard brightness level across all images. Lastly, images are clipped to 90% of their original size to eliminate boundary effects that may arise from the previous steps. This clipping ensures that only the most relevant portions of the image are retained for analysis, mitigating the influence of edge artifacts.

In our study, the method was further enhanced and applied in a new manner to reduce or eliminate the shadow effect caused by the circular patch. As a result, the angiographic images were more effectively enhanced. The modified method involves several steps, as depicted in **Figure 1.** Initially, Canny edge detection is applied to the original image, which identifies edges based on the maximum gradient magnitude in the ideal detector's plain estimate (Canny, 1986) Click or tap here to enter text.. The resulting mask from Canny edge detection is then inverted, and a custom 3×3 kernel is employed to refine the generated mask for the circular patch. Additionally, by carefully adjusting the erosion and dilation processes, the technique of morphological closing further improves the mask's quality. Finally, the mask is applied to the original image, and the Ben Graham enhancement technique is performed. The described approach ultimately enhances the performance of the Ben Graham image enhancement technique by eliminating the shadow from the circular patch.

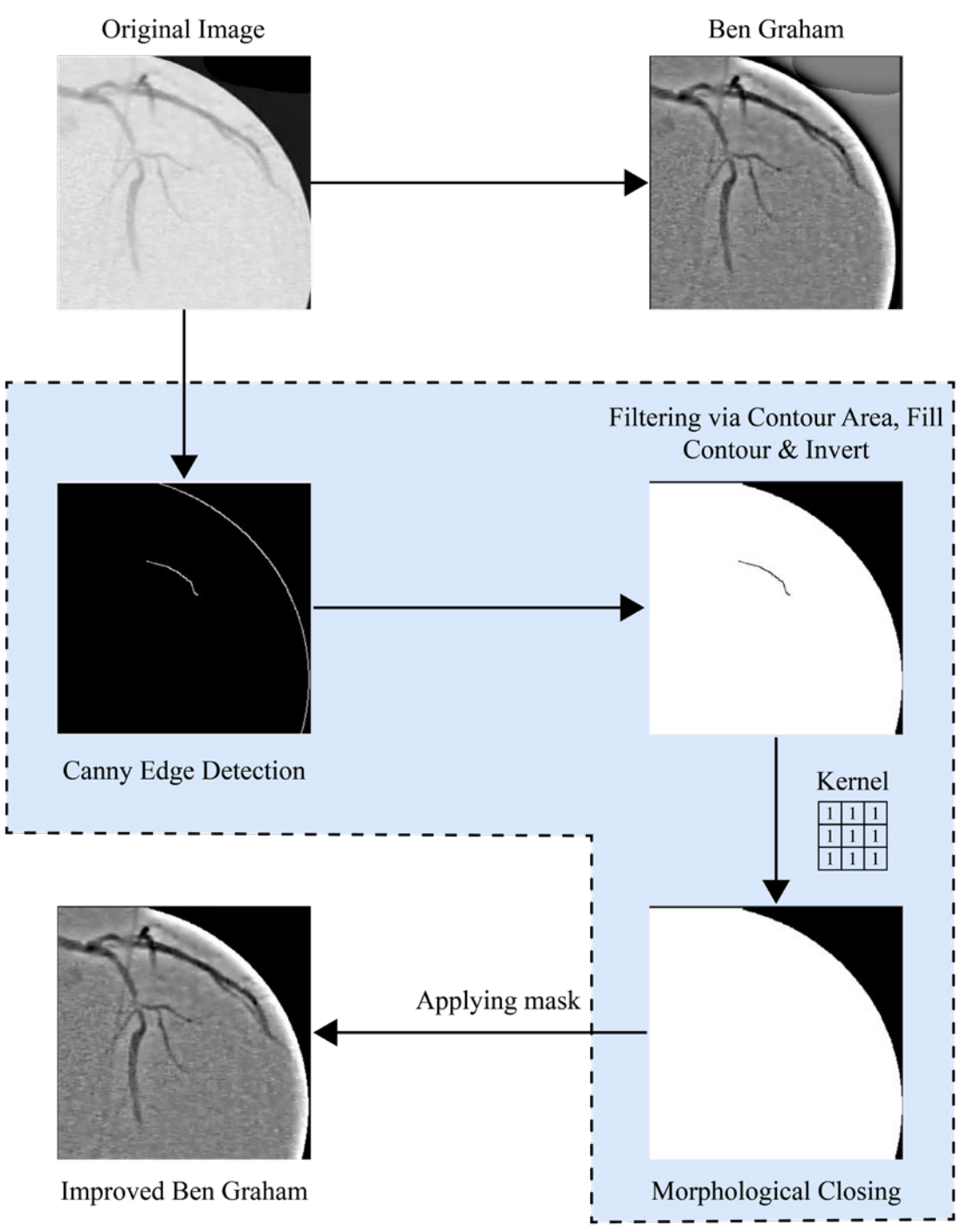

**Figure 1**. Workflow of the improved Ben Graham method for coronary angiogram image enhancement.

### 4.3 Multichannel Pre-processing

The strengths of CLAHE and the improved Ben Graham method were combined to maximize contrast enhancement in X-ray angiograms. CLAHE was applied to one channel to improve local contrast and vessel visibility, while the enhanced Ben Graham technique was applied to another channel to suppress background artifacts and emphasize vascular structures. These two channels were then concatenated to form a multichannel input image. Compared with unprocessed data or either technique applied independently, this representation achieved superior segmentation performance by providing complementary contrast features for the network. **Figure *2*** illustrates the preprocessing pipeline used to generate the multichannel angiogram images. **Figure 2** illustrates the preprocessing pipeline used to generate the multichannel angiogram images.

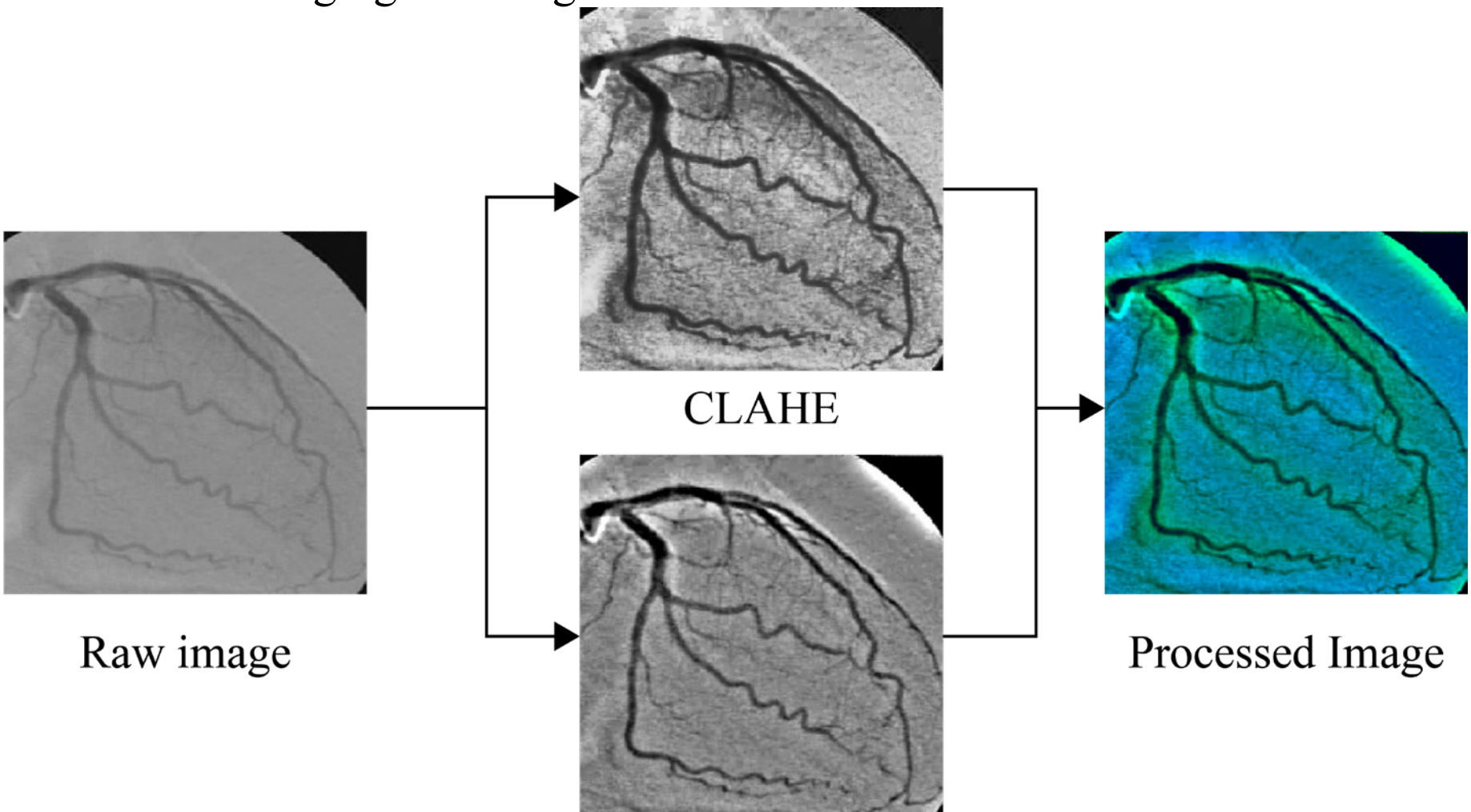

**Figure 2**. Multichannel image enhancement process using CLAHE and improved Ben Graham's method.

### 4.4 Data Augmentation

Data augmentation was employed to expand the training dataset and improve model generalization by introducing variability through transformations such as rotation, cropping, translation, flipping, and shifting. Deep neural networks are inherently tolerant to such transformations, making them suitable for learning complex vascular patterns in X-ray coronary angiograms. For the transfer learning-based convolutional neural network used in this work, relatively simple augmentations were sufficient to improve performance. These included rotations of ±5°, ±10°, ±15°, ±30°, ±45°, ±60°, ±75°, and ±90°, as well as translations of ±5% and ±10% in both vertical and horizontal directions, resulting in eight translation combinations (Samuel & Veeramalai, 2021). The augmentation expanded the training set by a factor of 20, yielding 5,550 images per fold, while test images were left unaltered to ensure unbiased evaluation.

### 4.5 Procedure Synopsis

All segmentation models were implemented using PyTorch, which consistently outperformed TensorFlow in preliminary experiments. The final models were developed with PyTorch 1.9 and Python 3.9, and the Self-ONN blocks were integrated using the FastONN library (Malik et al., 2020). To ensure consistency and comparability, identical training parameters were used across all networks. **Table *2*** summarizes the hyperparameters, which include batch size, learning rate, number of folds for cross-validation, learning rate drop factor, maximum epochs, early stopping criteria, loss function, optimizer, and the Self-ONN q-order.

**Table 2**. List of training parameters

| Training parameters | Parameter value |
|---|---|
| **Batch size** | 16 |
| **Learning rate** | 1e-4 |
| **Number of folds** | 5 |
| **Learning rate drop factor** | 0.2 |
| **Max epochs** | 200 |
| **Epoch patients** | 5 |
| **Early stopping criteria** | 20 |
| **Loss function** | Dice loss |
| **Optimizer** | Adam |

| Self-ONN q-order | 3 |
|---|---|

Overfitting was mitigated through early stopping: training was halted if no improvement in validation loss occurred within 20 epochs. Additionally, a learning rate scheduler reduced the rate by a factor of 0.2 if validation loss stagnated for 5 consecutive epochs. These measures ensured both efficient convergence and generalizable performance.

## 5. Methodology Analysis

The proposed pipeline for coronary artery segmentation from X-ray coronary angiograms (XCAs) consists of three stages, as illustrated in **Figure *3***. The first stage applies hybrid preprocessing to enhance vessel contrast, where CLAHE and the improved Ben Graham method are combined to generate multichannel inputs. The second stage performs segmentation using a Self-ONN–based UNet. In this step, ImageNet weights are applied to the encoder, while conventional Conv2D layers in the decoder are replaced with Self-ONN blocks, allowing the network to better capture the non-linear vessel geometry. This segmentation network produces an initial binary coronary artery mask.

The third stage refines the predicted mask through three investigated approaches. The first involves a deep learning–based refinement network, built on a ResNet50-UNet architecture, to suppress spurious false predictions. The second is a contour refinement module that removes small false-positive regions by identifying and inverting contours and applying bitwise operations with the original mask. The third is a patch line generation technique that reconnects disjoint vessel branches by detecting endpoints and generating patch lines, as detailed in Section 5.2. Together, these three stages form a comprehensive pipeline that enhances vessel continuity, reduces false predictions, and improves overall segmentation quality. The following sections present the datasets, preprocessing and augmentation strategies, model architecture, and refinement methods.

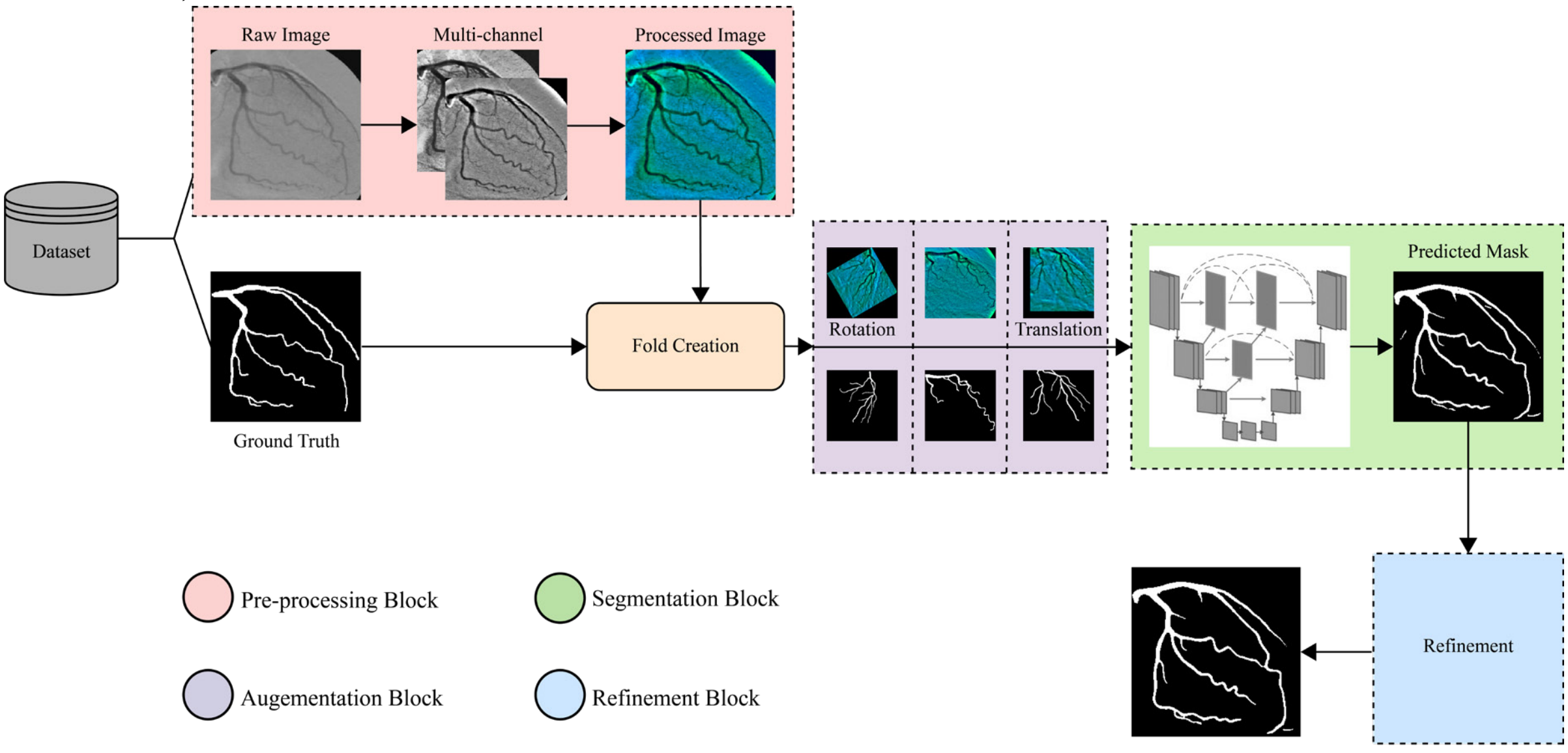

**Figure 3**. Overview of the proposed CASR-Net pipeline for coronary artery segmentation.

### 5.1 Segmentation Models

A range of encoder–decoder architectures was investigated for coronary artery segmentation from X-ray angiograms. These included UNet, UNet++, and variants with DenseNet, ResNet, Inception, EfficientNet, and Vision Transformer backbones. These networks follow the encoder–decoder paradigm, where the contracting path captures contextual information and the expanding path reconstructs vessel structures, with skip connections preserving spatial detail (Ronneberger et al., 2015). UNet++ further extends this design with nested convolutional blocks and deep supervision to improve feature representation (Stoyanov et al., 2018).

To optimize performance and reduce training time, transfer learning was applied using ImageNet-pretrained weights for all encoder backbones (Deng et al., 2009). The backbone variations explored included DenseNet121_UNet, DenseNet121_Unet++_scSE, DenseNet121_Linknet,

DenseNet201_Unet++, EfficientNet_b2_Unet++, Inception_v4_Unet++, and ResNet50_Self-ONN_Unet_q3. In addition to these, several recently proposed medical segmentation architectures were also evaluated, including ResNet50_MAnet (Fan et al., 2020), mit_b1_FPN (Li et al., 2021), MMDC-Net (Zhong et al., 2022), and TiM-Net (Hongbin Zhang et al., 2022), which incorporate multi-scale attention and transformer-based components. Furthermore, MedSAM, adapted from the Segment Anything Model, was included as a benchmark representing foundation-model approaches with strong generalization across medical imaging tasks.

This comprehensive set of models provided a robust baseline against which the performance of CASR-Net could be compared, covering conventional CNN-based networks, attention-augmented designs, transformer hybrids, and large-scale pretrained models.

### 5.1.1 Self-ONN

Multi-Layer Perceptrons (MLPs) and CNNs have a shared limitation: they rely on a homogenous network structure with linear neuron models, which inadequately reflects the diversity and complexity of biological neural systems. Generalized Operational Perceptrons (GOPs) and Operational Neural Networks (ONNs) address this by introducing heterogeneous and non-linear network models. GOPs, building on biological principles, demonstrate superior performance on complex tasks where traditional models fail. ONNs extend this approach, offering a more flexible framework by incorporating a diverse set of operators per neuron. ONNs adopt the fundamental concept from GOPs, thereby expanding beyond the exclusive reliance on linear convolutions within convolutional neurons by incorporating nodal and pool operators. These additions establish the operational layers and neurons while retaining two core constraints inherited from traditional CNNs: weight sharing and restricted (kernel-wise) connectivity. Illustrated in **Figure 4** are three operational layers and the kth neuron with 3x3 kernels within an ONN.

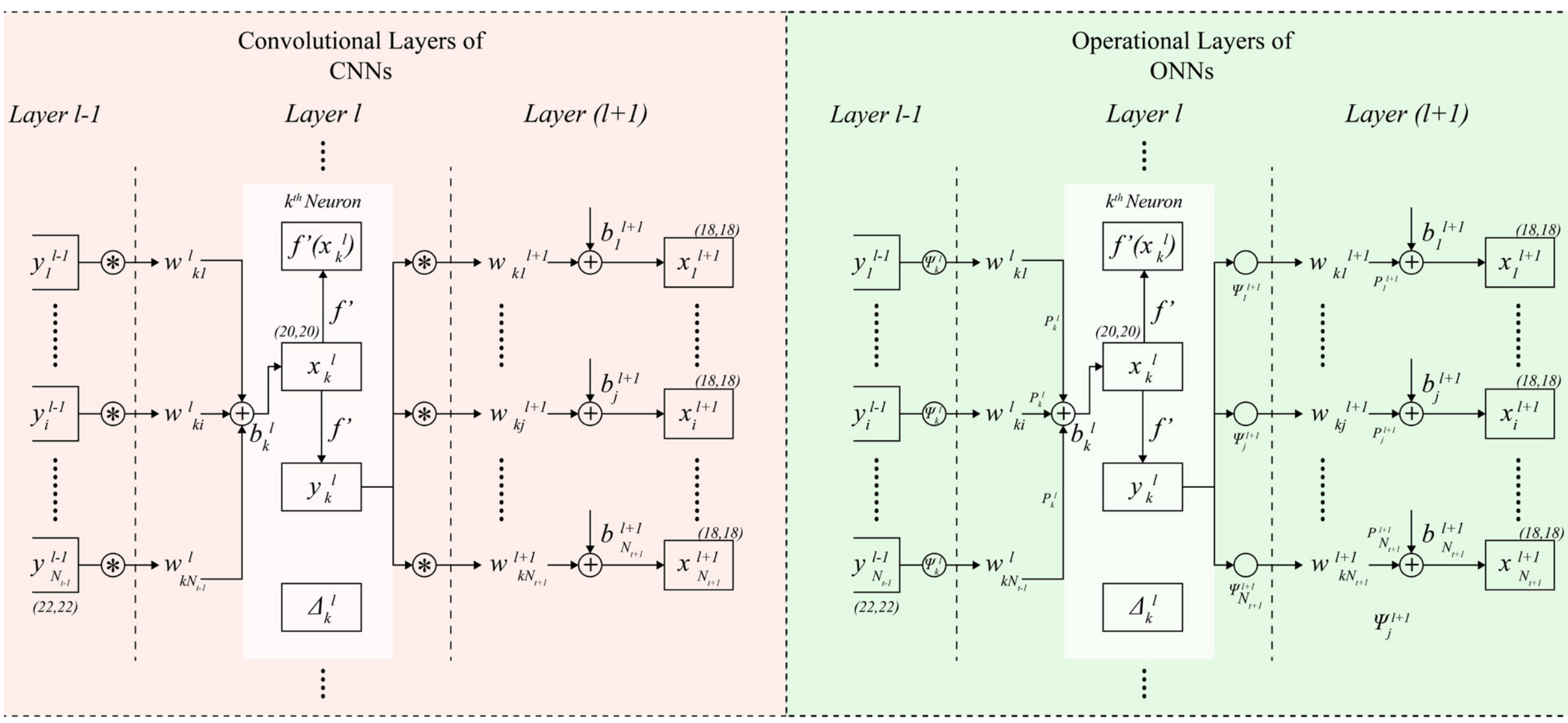

**Figure 4**: The illustration of the k-th neuron of a CNN (left) and an ONN (right)

The input map of the kth neuron in the current layer of an ONN, denoted as $x_k^l$, is derived by pooling the final output maps, $y^{l-1}_i$, of the preceding layer's neurons that have been operated on with their respective kernels, $w_{ki}^l$. It can be expressed in the following equation:

$$x_k^l = b_k^l + \sum_{i=1}^{N_{l-1}} \text{oper } 2D\left(w_{ki}^l, y_i^{l-1}, \quad '\text{NoZeroPad'}\right)$$

$$x_k^l(m,n)\Big|_{(0,0)}^{(M-1,N-1)} = b_k^l + \tag{3}$$

$$\sum_{i=1}^{N_{l-1}} \left( P_k^l \begin{bmatrix} \Psi_{ki}^l\left(w_{ki}^l(0,0), y_i^{l-1}(m,n)\right), \ldots, \\ \Psi_{ki}^l\left(w_{ki}^l(\mathrm{r,t}), y_i^{l-1}(m+r, n+t), \ldots\right), \ldots . \end{bmatrix} \right)$$

On the other hand, workings of a CNN layer can be expressed with the following equation:

$$x_k^l = b_k^l + \sum_{i=1}^{N_{l-1}} \text{conv } 2D\left(w_{ki}^l, y_i^{l-1}, \text{NoZeroPad}\right)$$
$$x_k^l(m,n)\Big|_{(0,0)}^{(M-1,N-1)} = \sum_{r=0}^{2} \sum_{t=0}^{2} \left(w_{ki}^l(r,t) y_i^{l-1}(m+r, n+t)\right) + \cdots \tag{4}$$

Self-ONNs build upon the principles of ONNs by introducing generative neurons that can adapt and optimize the nodal operator of each connection during the training process. This results in a higher level of heterogeneity and computational efficiency, as Self-ONNs do not require a predefined operator set library or an iterative search process to find the best operators. Instead, each neuron can create any combination of nodal operators, allowing for a more flexible and powerful modeling capability. The generative neurons in Self-ONNs use a composite nodal operator that can be iteratively created during backpropagation training without any restrictions. This composite operator can be expressed as a Q-th order Taylor approximation, where Q is the q-order of the polynomial. For example, the composite nodal function can be written as:

$$\Psi(\boldsymbol{w}, y) = w_0 + w_1 y + w_2 y^2 + \ldots + w_Q y^Q \tag{5}$$

During the forward propagation, the nodal operator for each kernel element is approximated by the composite nodal operator, resulting in a more diverse and optimized network. The forward propagation in Self-ONNs can be expressed as:

$$\Psi\left(\mathbf{w}, y_{(m+r,n+t)}\right) = w_{r,t,0} + w_{r,t,1} y_{(m+r,n+t)} + w_{r,t,2} y_{(m+r,n+t)}^2 + \ldots + w_{r,t,Q} y_{(m+r,n+t)}^Q \tag{6}$$

This allows each neuron to self-organize its nodal operators during training, optimizing the functions to maximize learning performance. Experimental results demonstrate that Self-ONNs outperform conventional ONNs and CNNs in terms of learning capability and computational efficiency, even with more compact networks (Kiranyaz et al., 2021)

### 5.1.2 Proposed Network

Self-ONN, introduced by (Kiranyaz et al., 2021) is a network designed to handle complex and nonlinear problems that traditional convolutional networks struggle with. Unlike operational neural networks (ONNs), which rely on a predefined library of operators and iterative search procedures, Self-ONNs employ generative neurons that can autonomously optimize their nodal operators during training. This self-organizing capability provides greater flexibility than conventional convolutional neural networks and eliminates the need for manually selecting operator sets (Malik et al., 2021).

Building on this principle, we propose a modified UNet architecture (**Figure 5**) with enhancements to both encoder and decoder. In the encoder, standard convolutional layers are replaced with DenseNet121 (excluding the final fully connected layer). Dense connections and feature reuse improve feature extraction, and ImageNet pre-trained weights accelerate convergence. Skip connections are maintained, linking encoder dense blocks to corresponding decoder layers, ensuring that the rich feature representations are preserved during reconstruction.

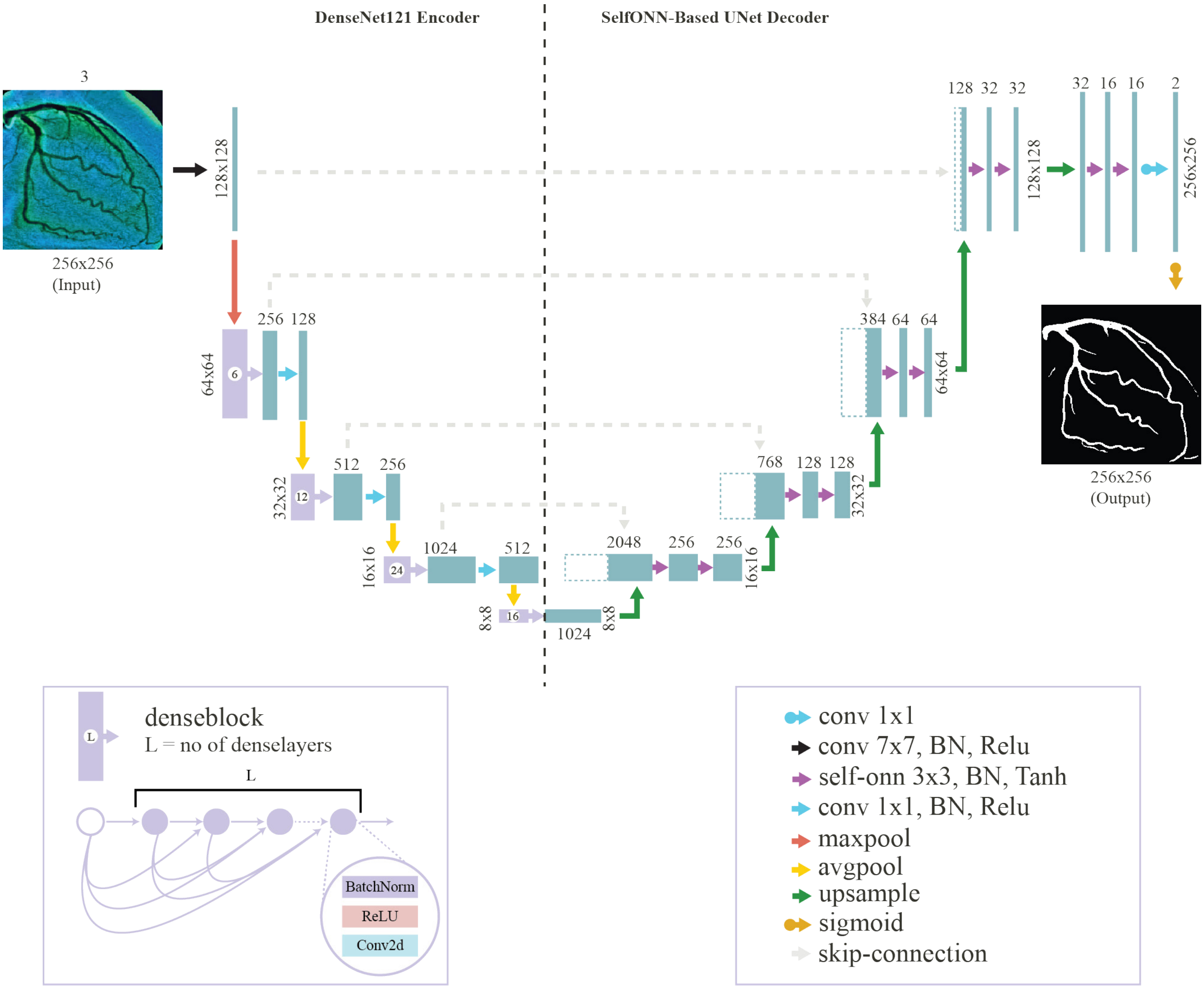

**Figure 5**. Architecture of the proposed segmentation network

In the decoder, conventional Conv2D blocks with ReLU activations are replaced by Self-ONN blocks with Tanh activation. This substitution allows the decoder to better capture complex vessel structures and maintains continuity in thin branches. Additionally, average pooling is used within the DenseNet encoder instead of max pooling, further supporting feature preservation. Together, these modifications enable the proposed network to more effectively segment coronary arteries, particularly in challenging cases with narrow or stenotic vessels.

### 5.2 Post-processing

Post-processing was applied to address two persistent challenges in coronary artery segmentation: (1) the presence of false positives, where background structures or noise are misclassified as vessels, and (2) discontinuities in vessel masks, particularly in low-contrast regions. To mitigate these issues, several refinement approaches were investigated, as outlined in the following subsections.

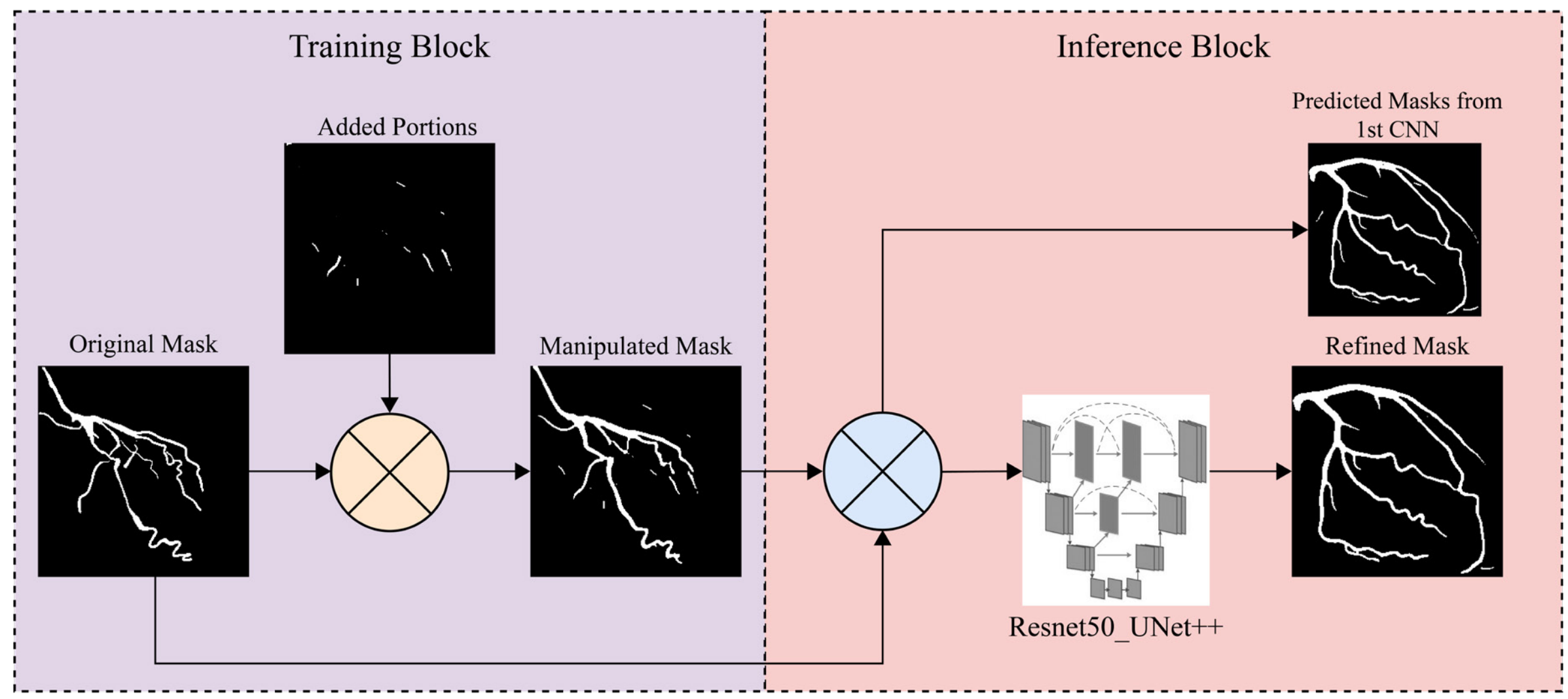

**Figure 6**. Deep learning-based refinement process for enhancing coronary artery segmentation.

### 5.2.1 Deep learning-based refinement

A refinement network based on UNet++ with a ResNet50 encoder was investigated to reduce false positives in coronary artery segmentation. To generate training data for this task, ground truth masks were deliberately modified by inserting small white clusters, simulating the spurious predictions observed in the DenseNet121–Self-ONN UNet outputs. Training on these augmented masks enabled the refinement network to learn to distinguish true vessel regions from erroneous fragments. The refinement workflow is illustrated in **Figure *6***.

ResNet50 was chosen as the encoder after comparing it with DenseNet, EfficientNet, and Vision Transformers. While all candidates support transfer learning, ResNet50 provided the most consistent balance of accuracy and efficiency. Its residual connections stabilized gradient flow, facilitated faster convergence, and preserved fine vessel structures, which was particularly beneficial for suppressing subtle false positives. EfficientNet and Vision Transformers demonstrated competitive accuracy in other vision tasks but required more computational resources and showed higher sensitivity to hyperparameter tuning. In contrast, ResNet50 yielded reliable performance across runs and has a strong record in biomedical imaging, supporting its adoption as the backbone for the UNet++ refinement model.

### 5.2.2 Contour refinement

A contour refinement strategy was investigated to remove small false-positive regions from the segmentation masks (**Figure 7**). The contour profile of the predicted mask was first extracted, and a heuristic area threshold was applied to identify contours corresponding to spurious detections. These small contours were inverted and combined with the original mask using a bitwise AND operation. This process effectively eliminated false positives and produced a refined mask with improved delineation of the coronary arteries.

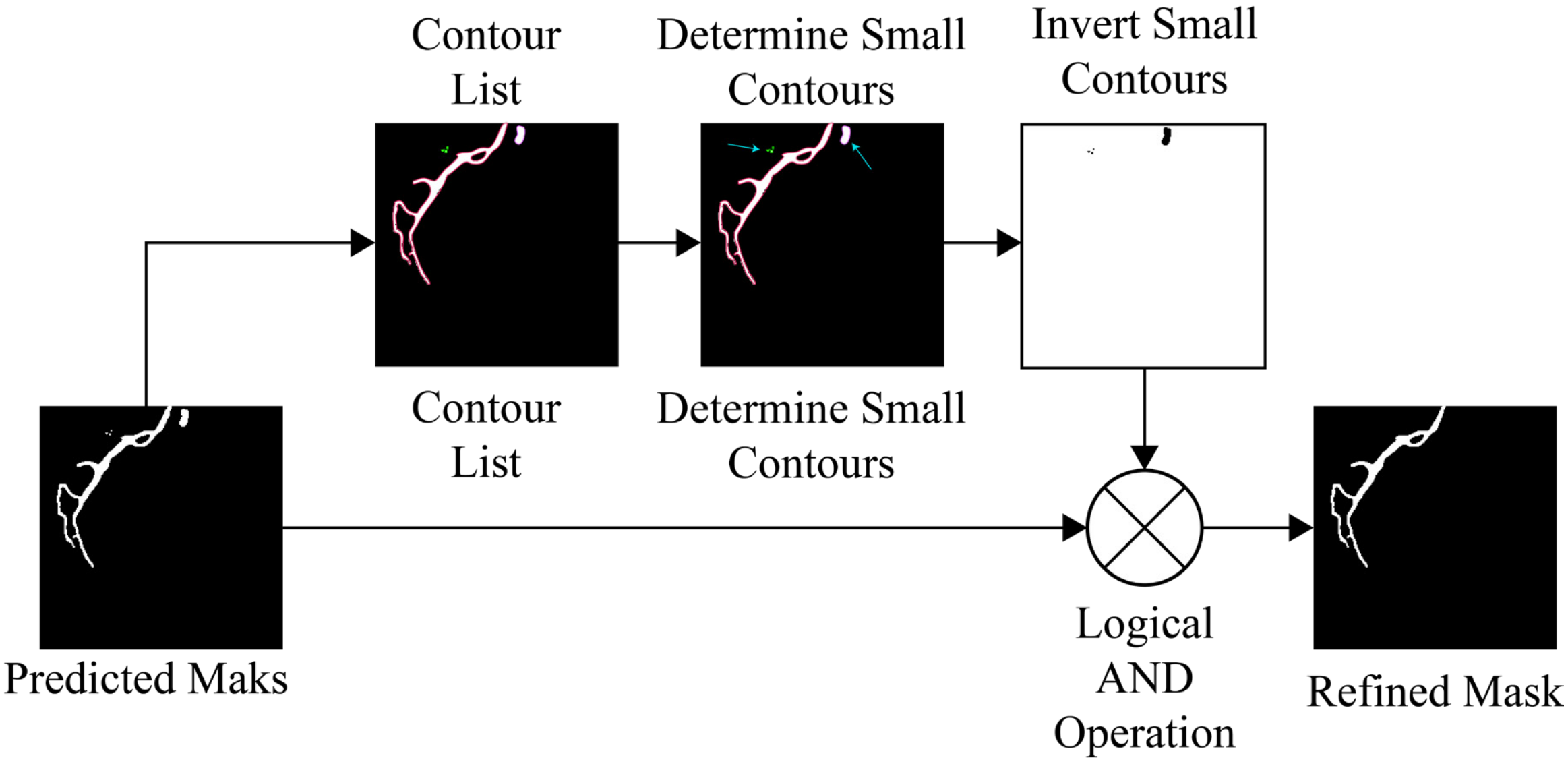

**Figure 7**. Workflow of the contour-based refinement process.

**5.2.3 Patch line generation**

Segmentation networks often struggle to preserve continuity in thin arterial branches, leading to fragmented or disjointed vessel masks. This issue is particularly evident in regions affected by stenosis, where narrow branches are more susceptible to breakage. To address this limitation, a patch line generation technique was investigated to restore continuity and minimize segmentation errors (**Figure 8**).

The process begins with morphological thinning, which reduces the predicted mask to a one-pixel-wide skeleton. A convolution kernel is then applied to detect branch endpoints, which are marked for potential reconnection. For each endpoint, the nearest neighboring endpoint is identified based on Euclidean distance. Candidate connections are then evaluated to determine whether they should be accepted as valid patch lines. To assess validity, a two-pixel-wide scanning region is created along the direction of the proposed connection. The pixel continuity within this region is compared with the candidate patch line. If the difference in pixel counts exceeds a predefined threshold, the connection is accepted as valid, as it indicates the patch line has successfully bridged a broken vessel segment. Connections that fall below the threshold are discarded as invalid. As shown in **Figure 9**, valid patch lines (highlighted in yellow) successfully reconnect fragmented branches, while invalid ones (highlighted in red) are excluded.

Finally, valid patch lines are merged with the original mask, producing an output where previously disconnected vessel segments are reconnected. This refinement improves the continuity of thin branches and yields more anatomically realistic segmentations, thereby enhancing both visual quality and quantitative performance metrics.

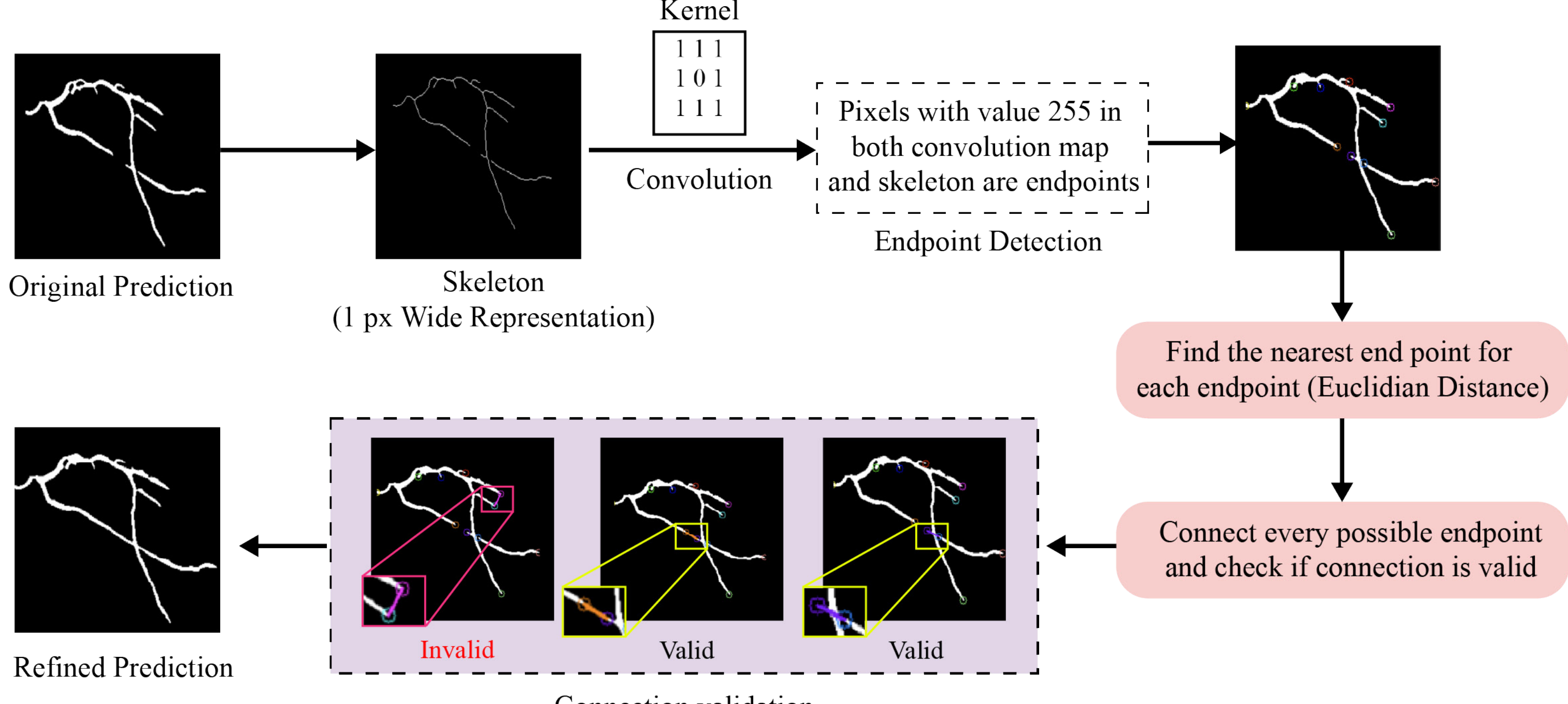

**Figure 8.** Connecting disjoint branches using endpoint detection and continuity checking.

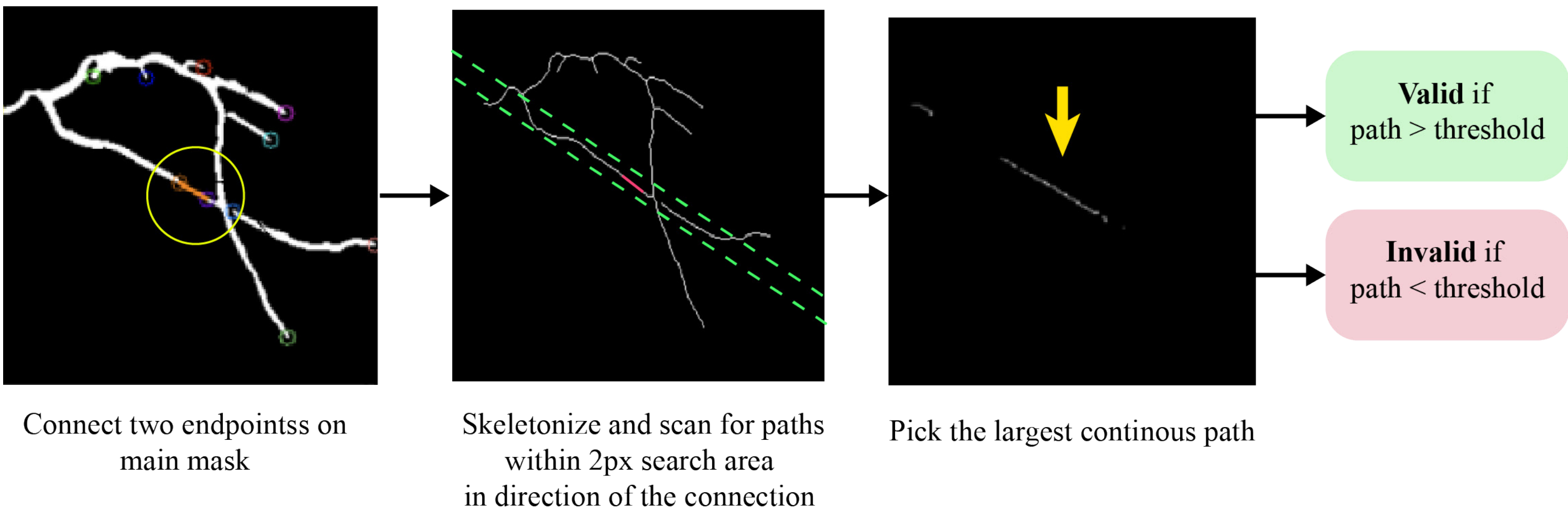

**Figure 9**. Detailed overview of the valid connection selection process.

### 5.3 Loss Function

Cross-entropy is a metric used to measure the disparity between two probability distributions for a given random variable or series of events. In classification tasks, the Binary Cross-Entropy (BCE) loss is commonly employed, and it proves to be effective for segmentation tasks as well. This is because semantic segmentation can be viewed as a pixel-level classification problem (Jadon, 2020). The BCE loss is defined by the following equation:

$$\text{Loss}_{(\text{BCE})} = \frac{1}{\text{N}} \sum_{i=0}^{N-1} -(y_i\, log(\hat{y}_i) + (1-y)\, log(1-\hat{y}_i)) \tag{7}$$

When ground truth data is accessible, the evaluation of segmentation performance often relies on the DSC, a metric that quantifies the overlap between predicted masks and the actual ground truth. The DSC is commonly employed for segmentation tasks to compute the similarity index between the anticipated masks and the ground truth (Zhao et al., 2020). To express the dice loss, the following equation is utilized:

$$\text{Loss}_{(\text{DICE})} = 1 - \frac{\sum_{i=0}^{N-1} \hat{y}_i y_i}{\sum_{i=0}^{N-1} y_i^2 + \sum_{i=0}^{N-1} \hat{y}_i^2 + \epsilon} \quad (8)$$

In Equations (7) and (8), N represents the total number of pixels, $y_i$ corresponds to the $i^{th}$ pixel in the ground truth mask, and $\hat{y}_i$ represents the ith pixel for the predicted mask. For the training and investigation of segmentation models in this study, both of the mentioned loss functions were utilized, along with a hybrid one that combined both. However, for the in-depth examination, Dice loss was predominantly used, as the initial investigation indicated its superior performance compared to BCE loss.

### 5.4 Evaluation Matrices

To evaluate the proposed techniques for coronary artery segmentation, quantitative assessments were conducted. The positive class was defined as the coronary arteries (foreground), while the background represented the negative class. Results from all folds of the 5-fold cross-validation were combined to compute the metrics.

The primary evaluation metrics were Accuracy (Acc), Intersection over Union (IoU), and Dice Similarity Coefficient (DSC), which are widely used in the literature for medical image segmentation (Kim et al., 2022; Li et al., 2022; Liu, Yang, Tian, Cao, et al., 2022; Qiblawey et al., 2021)(Kim et al., 2022; Liu et al., 2022; Li et al., 2022; Qiblawey et al., 2021). Their definitions are as follows:

$$Accuracy = \frac{TP + TN}{TP + TN + FP + FN} \quad (9)$$

$$Intersection\ over\ Union\ (IoU) = \frac{TP}{TP + FP + FN} \quad (10)$$

or

$$Intersection\ over\ Union\ (IoU) = \frac{Area\ of\ Overlap}{Area\ of\ Union} \quad \text{[Pixel area wise]}$$

$$Dice\ Similarity\ Coefficient\ (DSC) = \frac{2TP}{2TP + FP + FN} \quad (11)$$

or

$$Dice\ Similarity\ Coefficient\ (DSC) = \frac{2 \times Area\ of\ Overlap}{Total\ Area} \quad \text{[Pixel area wise]}$$

$$False\ Positive\ Rate\ (FPR) = \frac{FP}{FP + TN} \quad (12)$$

$$False\ Negative\ Rate\ (FNR) = \frac{FN}{FN + TP} \quad (13)$$

$$Precision\ (P) = \frac{TP}{TP + FP} \quad (14)$$

$$Sensitivity\ (SN) = \frac{TP}{TP + FN} \quad (15)$$

$$Specificity\ (SP) = \frac{TN}{TN + FP} \quad (16)$$

Finally, to better capture vessel continuity, the clDice score was also employed. Unlike IoU and DSC, which primarily measure pixel-level overlap, clDice incorporates topological information by evaluating overlap between the centerlines of predicted and ground truth vessels (Shit et al., 2021). This metric is particularly important for assessing coronary artery segmentation, as it directly reflects the ability of a model to preserve connectivity in thin branches.

## 6. Result and Discussion

In this section, the findings of the study are described in three sub-sections. The first sub-section discusses the impact of image pre-processing and the choice of loss function on the segmentation performance. The second sub-section presents a comparative analysis of various networks, including the proposed network. The third sub-section details the advantages of Self-ONN, and the fourth sub-section outlines the refinement module's role and evaluates the overall performance of the proposed coronary artery segmentation pipeline applied to X-ray angiogram images. The fifth subsection reports a detailed ablation study on our proposed model and lastly, the sixth sub-section provides a comparative analysis of our proposed method with existing approaches.

### 6.1 Effect of Image Pre-processing for CA Segmentation

**Table 3** summarizes the segmentation performance based on various image pre-processing techniques. The investigation was conducted on the combined dataset using 5-fold cross-validation, and the average results are reported. Established segmentation networks like InceptionV4_UNet++, Attention_Super_FPN, and DenseNet201_UNet++ were evaluated, but the proposed network outperformed them and is therefore presented in this section. The study explored four classes of data: original images, images processed with the CLAHE technique, images processed with the improved Ben Graham algorithm, and images processed using the proposed multichannel technique combining features from the previous two. Initially, experiments were performed using both TensorFlow and PyTorch frameworks, with the investigation in the PyTorch framework showing better results across all modules, leading to the study's continuation in PyTorch. Three types of loss functions, namely BCE, DICE, and hybrid loss functions, were tested, as described in Section 4.8. Although the BCE loss function is commonly used in segmentation tasks, the DICE loss function outperformed both BCE and the hybrid loss function in this case. Across all processed classes, the DICE loss function consistently yielded superior results. Notably, the proposed multichannel pre-processing technique surpassed the other methods significantly, achieving a Dice Similarity Coefficient (DSC) of 75.97%. This performance is considerably higher than the original dataset, which produced a DSC of only 75.12% over the same framework. The results indicate that the proposed multichannel image pre-processing technique substantially improves the network's performance by combining contrast enhancement features from two distinct algorithms.

**Table 3.** Performance evaluation (in %) of different image pre-processing techniques for coronary artery segmentation

| Pre-processing | Evaluation Matrix | | | | | | | |
|---|---|---|---|---|---|---|---|---|
| | Acc | IoU | DSC | P | SN | SP | FNR | FPR |
| **Original** | 97.45 | 60.16 | 75.12 | 74.02 | 76.45 | 98.56 | 23.55 | 1.44 |
| **CLAHE** | 97.42 | 60.10 | 75.08 | 73.21 | 77.30 | 98.48 | 22.70 | 1.52 |
| **Improved Ben Graham** | 97.47 | 60.86 | 75.66 | 73.33 | 78.18 | 98.49 | 21.82 | 1.51 |
| **multichannel** | 97.55 | 61.26 | 75.97 | 75.08 | 77.06 | 98.68 | 23.05 | 1.36 |

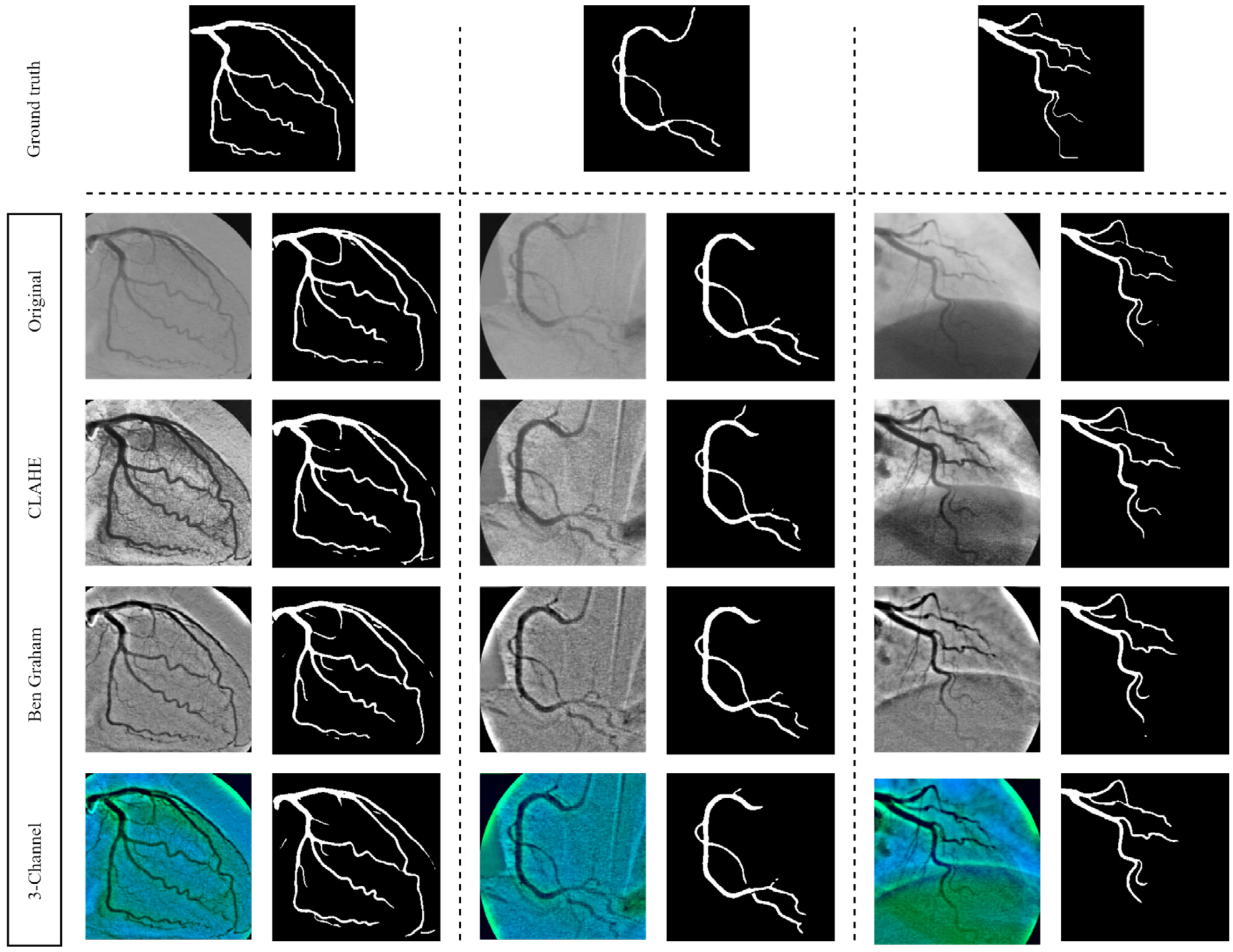

**Figure 10**. Comparative analysis of various pre-processing approaches for coronary artery segmentation.

**Figure 10** illustrates a comparative demonstration of segmentation performance achieved by adopting various image processing techniques. The attached samples clearly indicate that the proposed multichannel approach outperforms other techniques in finely segmenting the coronary arteries (CAs).

### 6.2 Segmentation Performance

**Table 4** presents a summary of performance metrics for the investigated networks. The experiment utilized a 5-fold cross-validation approach, and the average results from the folds are tabulated. Quantitatively, the proposed network demonstrates superior performance with a DSC of 76.10 and IoU of 61.43, outperforming all conventional segmentation networks consistently. It also demonstrates a reasonably fast average per image inference time at 0.696 seconds. The proposed refinement process yields higher-scoring matrices and improved vessel features compared to traditional methods, as evidenced by quantitative analysis. However, it may occasionally result in the unintended removal of certain branches. Despite this limitation, the proposed approach generally performs better overall, albeit with some reduction in qualitative performance. The UNet decoder-based networks achieved higher scores in all the scoring matrices. In addition to these global metrics, we further evaluated topology-aware performance using clDice, which better reflects vessel continuity. The proposed network achieved the highest clDice score of 79.38, closely followed by Densenet121_Unet++_scSE (79.26) and DenseNet201_UNet++ (79.25). This confirms that the proposed approach not only improves global metrics but also preserves thin vessel structures effectively.

**Table 4**. Performance evaluation of different networks using 5-fold cross-validation for coronary artery segmentation

| Pre-processing | Network | Trainable Parameters (Million) | Average Inference time (s) ▼ | Acc ▲ | IoU ▲ | DSC ▲ | P▲ | SN▲ | SP▲ | FNR ▼ | FPR ▼ |
|---|---|---|---|---|---|---|---|---|---|---|---|
| Proposed (3-Ch) | **Proposed network** | 26.9 | 0.696 | **97.52** | **61.43** | **76.10** | **73.79** | **78.61** | **98.52** | **21.39** | **1.48** |
| | DenseNet121_UNet | 13.6 | 0.681 | 97.54 | 60.67 | 75.51 | 75.74 | 75.33 | 98.72 | 24.66 | 1.28 |
| | **Densenet121_Unet++_scSE** | 14.2 | 1.309 | **97.48** | **61.18** | **75.92** | **74.52** | **77.36** | **98.58** | **22.58** | **1.38** |
| | Densenet121_Linknet | 10.4 | 0.702 | 97.49 | 60.51 | 75.39 | 74.32 | 76.52 | 98.60 | 23.48 | 1.40 |
| | **DenseNet201_UNet++** | 48.6 | 1.144 | **97.55** | **61.12** | **75.87** | **75.31** | **76.52** | **98.67** | **23.48** | **1.33** |
| | Efficientnet_b2_Unet++ | 10.4 | 0.640 | 97.48 | 60.35 | 75.27 | 74.38 | 76.24 | 98.60 | 23.76 | 1.40 |
| | Inception_v4_Unet++ | 59.3 | 0.769 | 97.51 | 60.84 | 75.65 | 74.56 | 76.84 | 98.60 | 23.16 | 1.40 |
| | Resnet50_Self-ONN_Unetq3 | 50.53 | 0.402 | 97.52 | 60.51 | 75.40 | 75.52 | 75.44 | 98.69 | 24.56 | 1.31 |
| | Resnet50_MAnet | 147.4 | 0.488 | 97.53 | 60.47 | 75.37 | 75.63 | 75.17 | 98.71 | 24.83 | 1.29 |
| | mit_b1_FPN | 15.03 | 0.632 | 97.27 | 58.73 | 74.00 | 71.06 | 77.24 | 98.33 | 22.76 | 1.67 |
| | MMDC-Net | 19.36 | 1.554 | 97.00 | 60.95 | 75.39 | 65.32 | 89.91 | 97.37 | 10.09 | 2.63 |
| | TiM-Net | 104.06 | 1.834 | 96.97 | 43.73 | 56.52 | 84.50 | 48.71 | 99.52 | 51.29 | 0.48 |
| | MedSAM | 93.7 | 1.512 | 97.43 | 58.20 | 73.56 | 76.12 | 71.40 | 98.80 | 28.60 | 1.20 |
| CLAHE | Proposed network | 26.9 | 0.692 | 97.35 | 60.04 | 75.03 | 71.77 | 78.86 | 98.33 | 21.14 | 1.67 |
| Ben Graham | Proposed network | 26.9 | 0.694 | 97.41 | 60.77 | 75.59 | 71.87 | 79.77 | 98.34 | 20.23 | 1.66 |
| N/A | Proposed network | 26.9 | 0.698 | 97.40 | 60.18 | 75.14 | 72.56 | 78.11 | 98.42 | 21.89 | 1.58 |

Interestingly, despite having a shallower architecture, DenseNet121 outperforms DenseNet201 slightly. **Figure 11** provides a qualitative comparison of the segmented results from the three best-performing networks, revealing that the proposed network excels in segmenting smaller branches due to the utilization of the Self-ONN block's image restoration feature. Taken together, the top three performing networks for this dataset are the proposed network, Densenet121_Unet++_scSE, and DenseNet201_UNet++, (Roy et al., 2018 based on DSC, IoU, and clDice.

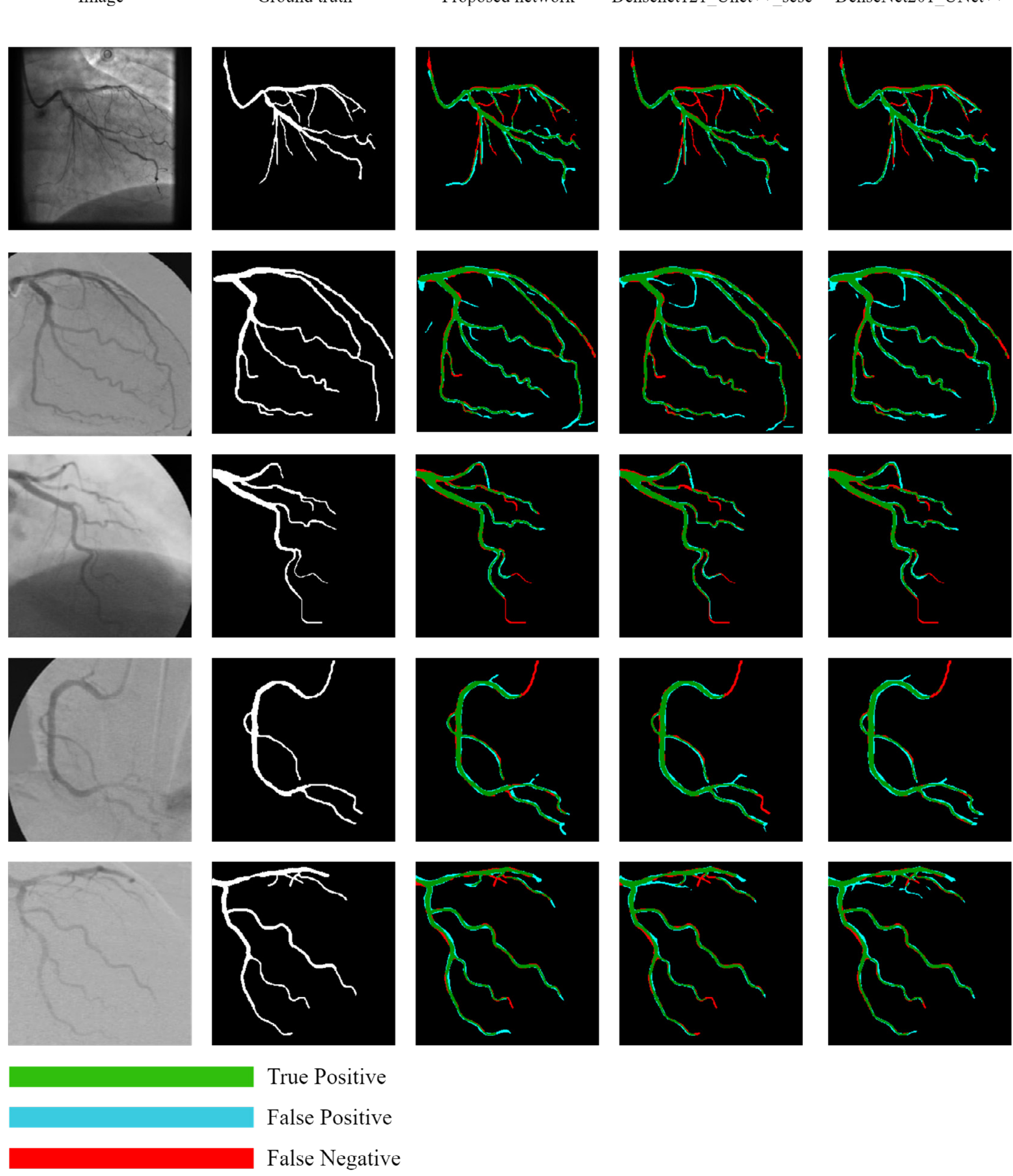

**Figure 11.** Sample predicted masks highlighting true positives (green), false positives (cyan), and false negatives (red) from the top 3 networks.

In segmentation studies, DSC and IoU are crucial evaluation metrics as both IoU and DSC directly reflect the accuracy of the segmentation in terms of spatial overlap, making them more relevant for tasks where the correct delineation of objects is critical. **Table 4** provides a comprehensive summary of the 5-fold cross-validation results for coronary artery segmentation, comparing our proposed network with various other models. Each evaluation metric is annotated with an arrow to indicate whether larger or smaller values are preferable. The proposed network achieves an IoU of 61.43 and a DSC of 76.10, surpassing the other top models in these critical metrics, thereby demonstrating its superior segmentation capability. While the Resnet50_MAnet model has a marginally faster inference time (approximately 200ms less), these advantages do not outweigh the near 1% IoU improvement and the consistently high performance demonstrated by the proposed network. The proposed network's relatively lower parameter count further enhances its memory efficiency, making it more practical for deployment in resource-constrained environments.

Additionally, the proposed network incorporates a refinement module that significantly reduces falsely classified regions, a common issue in conventional networks that lack such refinement blocks.

The integration of the Self-ONN within the decoder block enhances the network's robustness and adaptability, enabling more accurate segmentation even in challenging datasets.

### 6.3 Refinement Block Performance

**Figure** *12* compares the three refinement techniques investigated in this study, illustrating their ability to improve segmentation across diverse cases. **Figure** *13* further highlights the differences between unrefined and refined outputs. Even strong baseline models, such as DenseNet201, frequently produce false positives or capture irrelevant regions in the predicted masks. The refinement approaches, however, effectively suppress these artifacts, yielding cleaner and more reliable vessel boundaries. Conventional segmentation networks also tend to generate fragmented masks, particularly in narrow vessel regions where continuity is difficult to maintain. The patch line generation technique addresses this limitation by reconnecting disjoint branches with synthetic patch lines, thereby improving vessel completeness. When combined with contour refinement, which removes small false-positive regions, these two strategies significantly improve mask quality. Together, they provide notable gains in both visual clarity and quantitative metrics compared to networks without post-processing. It should be noted, however, that the refinement block is not universally optimal. Its effectiveness depends on the specific characteristics of the input images. For instance, patch line generation is most beneficial in cases where branch discontinuities are prominent, while contour refinement is particularly effective in suppressing spurious peripheral regions. Consequently, refinement should be viewed as a targeted post-processing step that enhances segmentation in cases with vessel breakages or excessive false positives, rather than as a one-size-fits-all solution.

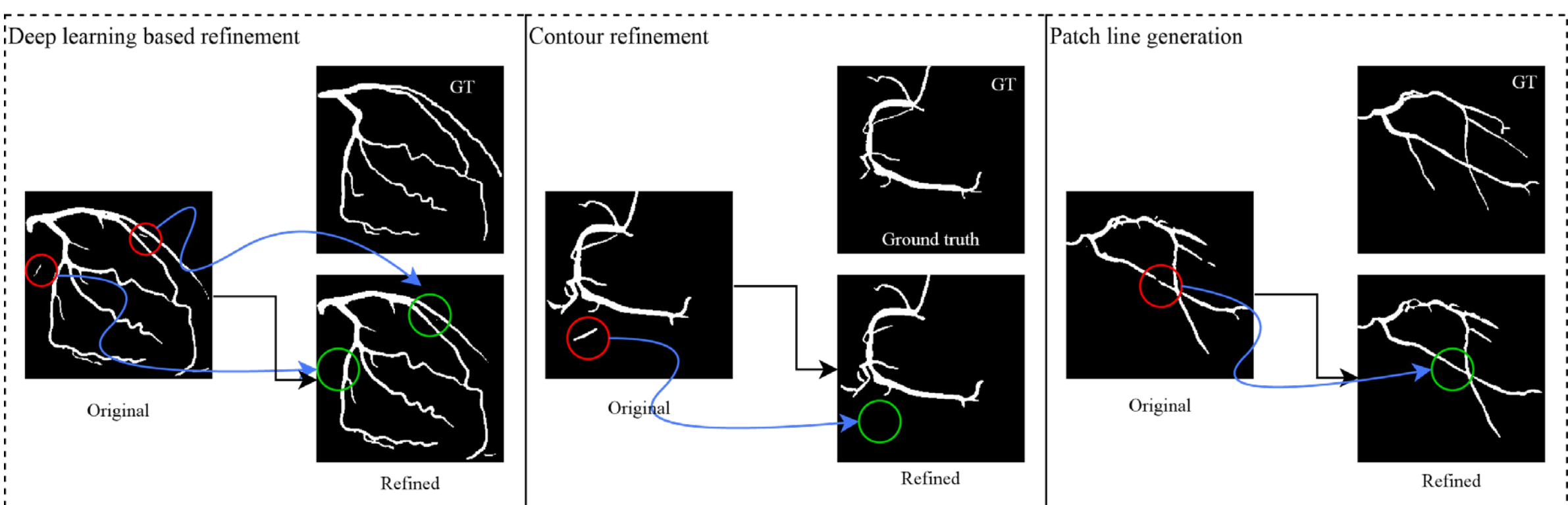

**Figure 12.** Comparison of the three refinement techniques: deep learning-based, contour refinement, and patch line generation.

While visual inspection suggests improvements, a quantitative analysis of the trade-offs reported in

**Table 5** provides a clearer picture of each refinement method's impact. Contour refinement demonstrates the most favorable balance, reducing the FNR by 1.66 pixels with only a negligible 0.12-pixel increase in FPR, resulting in the highest DSC of 76.10% and improved vessel continuity. In contrast, the deep learning–based refinement produces an unfavorable trade-off, as its slight reduction in FPR comes at the cost of a 1.36-pixel increase in FNR, reflecting erosion of true vessel segments and a decline in DSC. Patch line generation, meanwhile, shows a largely neutral effect: while it reconnects, vessel fragments and decreases FNR, this benefit is offset by an increase in FPR due to erroneous connections, leading to negligible improvement in overall performance. These findings highlight that refinement is best regarded as a targeted strategy, with contour refinement proving most effective for continuity, while the other methods show either limited or offsetting benefits.

**Table 5.** Performance comparison of post-processing techniques for coronary artery segmentation refinement

| Post-processing | Inference time (s) ▼ | Acc ▲ | IoU ▲ | DSC ▲ | P ▲ | SN ▲ | SP ▲ | FNR ▼ | FPR ▼ |
|---|---|---|---|---|---|---|---|---|---|
| N/A | 0.696 | 97.55 | 61.26 | 75.97 | 75.08 | 77.06 | 98.68 | 23.05 | 1.36 |
| Contour refinement | 0.6961 | 97.52 | 61.43 | 76.10 | 73.79 | 78.61 | 98.52 | 21.39 | 1.48 |
| DL based refinement | 1.396 | 97.51 | 60.45 | 75.34 | 75.19 | 75.59 | 98.67 | 24.41 | 1.33 |
| Patch line generation | 0.6986 | 97.51 | 60.98 | 75.75 | 74.38 | 77.24 | 98.58 | 22.76 | 1.42 |

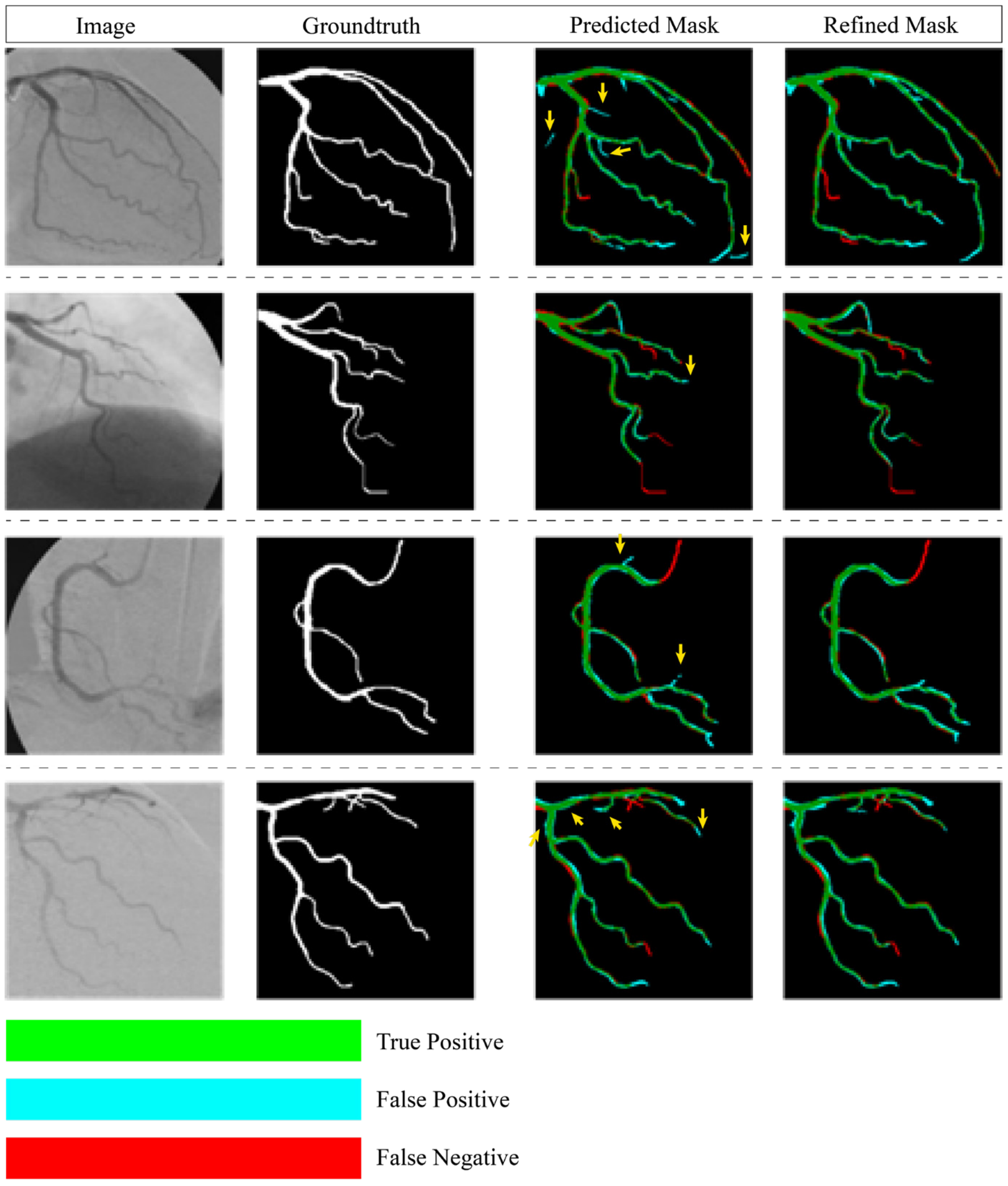

**Figure 13.** Comparison of predicted masks before and after refinement for our proposed model. (yellow marked areas show a decrease in false positives)

While the proposed approach achieves promising results in segmenting coronary arteries XCA images, several limitations must be acknowledged. First, the model has not been validated on external datasets, and its generalizability to images acquired under different conditions remains uncertain. Second, the computational demand and processing time are relatively high, suggesting that further optimization of the network architecture is necessary for more efficient deployment. Finally, the preprocessing and post-processing techniques, though effective in many cases, are partly heuristic and may not consistently improve results across all scenarios. Developing more adaptive and generalized solutions would be critical to ensuring the robustness and clinical applicability of future coronary artery segmentation frameworks.

### 6.4 Ablation Study

This section details the ablation studies conducted to assess the individual contributions of each component in our proposed model. All reported results in this section are on the validation set using 5-fold cross validation. This section is divided into 3 subsections discussing impact of different submodules, q-order values and loss functions on our proposed model.

#### 6.5.1 Impact Assessment of Different Submodules

This ablation study evaluates the contribution of each submodule in the proposed network, as summarized in **Table 6**. Beginning with the baseline UNet, the integration of DenseNet121 provided substantial gains across the primary evaluation metrics. Adding the Self-ONN module further improved performance, particularly in preserving vessel continuity and reducing segmentation errors. In contrast, incorporating the scSE module negatively impacted results by increasing false positives, which lowered precision, IoU, and DSC, despite a slight improvement in sensitivity. These findings indicate that the combination of UNet with DenseNet121 and Self-ONN offers the most effective balance of accuracy and robustness, and this configuration was adopted as the final model

**Table 6.** Performance evaluation of different submodules of our proposed segmentation network

| Model | Acc▲ | IoU▲ | DSC ▲ | P▲ | SN▲ | SP▲ | FNR ▼ | FPR ▼ |
|---|---|---|---|---|---|---|---|---|
| UNet | 97.50 | 58.44 | 73.77 | 75.08 | 72.63 | 98.77 | 27.37 | 1.23 |
| UNet + DenseNet121 | 97.57 | 60.48 | 75.37 | 74.07 | 76.74 | 98.63 | 23.26 | 1.37 |
| UNet + DenseNet121 + SelfONN | 97.57 | 60.96 | 75.74 | 73.42 | 78.23 | 98.56 | 21.77 | 1.44 |
| UNet + DenseNet121 + SelfONN + scSE | 97.47 | 60.11 | 75.09 | 71.82 | 78.81 | 98.42 | 21.19 | 1.58 |

#### 6.5.2 Performance Variation with Different q-order Values

This subsection examined the effect of different q-order values in the Self-ONN modules on model performance, as summarized in Table 7. A q-order of 3 produced the best overall results, achieving the highest IoU and DSC while maintaining a balanced trade-off between precision, sensitivity, and specificity. Increasing the q-order to 5 slightly reduced IoU and DSC but yielded a marginal gain in precision, reflecting a shift toward identifying true positives at the cost of overall segmentation quality. At q = 7, performance declined more sharply, with lower IoU and DSC, a modest precision gain, but reduced sensitivity and increased FNR, indicating missed true vessel segments. These findings suggest that lower q-order values provide more stable and accurate performance, with q = 3 offering the optimal balance across metrics. Consequently, q = 3 was selected for the final model configuration.

**Table 7.** Comparative performance evaluation of different q-values for our proposed network

| q-order | Acc▲ | IoU▲ | DSC▲ | P▲ | SN▲ | SP▲ | FNR▼ | FPR▼ |
|---|---|---|---|---|---|---|---|---|

| 3 | 97.57 | 60.96 | 75.74 | 73.42 | 78.23 | 98.56 | 21.77 | 1.44 |
|---|---|---|---|---|---|---|---|---|
| 5 | 97.55 | 59.64 | 74.72 | 74.67 | 74.82 | 98.70 | 25.18 | 1.30 |
| 7 | 97.48 | 57.07 | 72.67 | 77.55 | 68.36 | 98.98 | 31.64 | 1.02 |

### 6.5.3 Comparative Analysis of Loss Functions

This ablation study evaluated the effect of different loss functions on model performance, as shown in Table 8. BCE achieved an accuracy of 96.36% and a DSC of 70.44%, with the highest sensitivity (89.43%), but its lower precision and higher FPR limited overall effectiveness. Dice loss delivered the strongest results, with an accuracy of 97.57%, the highest IoU (60.96%), and a DSC of 75.74%, providing the best balance between precision and sensitivity. The compound loss, defined as the average of BCE and Dice, produced the weakest outcomes, with the lowest accuracy and IoU and only moderate DSC. These findings indicate that Dice loss offers the most reliable trade-off across evaluation metrics, making it the preferred choice for the proposed model.

**Table 8.** Performance comparison of using different loss functions of our proposed model

| Loss Function | Acc▲ | IoU▲ | DSC▲ | P▲ | SN▲ | SP▲ | FNR▼ | FPR▼ |
|---|---|---|---|---|---|---|---|---|
| Binary Cross-Entropy Loss | 96.36 | 54.38 | 70.44 | 58.15 | 89.43 | 96.71 | 10.57 | 3.29 |
| Dice Loss | 97.57 | 60.96 | 75.74 | 73.42 | 78.23 | 98.56 | 21.77 | 1.44 |
| Compound Loss | 96.26 | 53.79 | 69.95 | 57.76 | 88.67 | 96.65 | 11.33 | 3.35 |

### 6.6 Performance Comparison with Existing Approaches

Table 9 presents a comparative summary of recent state-of-the-art methods for coronary artery segmentation alongside the proposed approach. These models reflect diverse architectural strategies. DBCU-Net combines U-Net and DenseNet with bidirectional ConvLSTM layers to capture spatial–temporal features but struggles with small distal vessels and risks overfitting due to limited training data. SE-RegUNet incorporates squeeze-and-excitation blocks with RegNet encoders to recalibrate features dynamically, offering efficiency gains. ASCARIS employs an attention-based nested U-Net to strengthen feature extraction but often fails to maintain continuity at bifurcations. CIDN introduces Bio-inspired Attention and Multi-scale Interactive Blocks to enhance contextual feature learning, though its performance declines in low-contrast settings. Residual-Attention UNet++ integrates residual and attention mechanisms into UNet++ for improved stability and finer detail capture, but its complexity increases computational cost.

**Table 9** Comparing our proposed method with existing state-of-the-art approaches

| Reference | Test Dataset | Approach | Performance |
|---|---|---|---|
| **(Chang et al., 2024)** | Private dataset (109 images)<br>DCA1 dataset (134 images) | SE-RegUNet 4GF | Dice Similarity Coefficient - 0.7217 (Private dataset)<br>Dice Similarity Coefficient - 0.7621 (DCA1 dataset) |
| **(Zhang et al., 2024)** | JMA dataset (100 images) | Context Interactive Deep Network (CIDN) | Accuracy - 0.9757 (JMA dataset)<br>Accuracy - 0.9795 |

| | DCA1 dataset (134 images) | | (DCA1 dataset) |
|---|---|---|---|
| **(Shen et al., 2023)** | Private dataset (20 images) | DBCU-Net | Accuracy - 0.9850 |
| **(Algarni et al., 2022)** | DCA1 dataset (30images) | ASCARIS model based on Attention based Nested UNet | Accuracy - 0.9700 |
| **(Li et al., 2022)** | DCA1 dataset (30images) | Residual-Attention UNet++ | Dice Similarity Coefficient - 0.7248 Intersection over Union - 0.6657 |
| **Proposed** | **DCA1 dataset + Angiographic Dataset for Stenosis Detection (70 images per fold with 5-fold-cross validation)** | **CASR-Net** | **Dice Similarity Coefficient - 0.7610 Intersection over Union - 0.6143 Accuracy – 0.9752 clDice-0.7936** |

The proposed CASR-Net advances beyond these approaches by integrating a Self-ONN decoder with a multichannel preprocessing pipeline and targeted refinement strategies to achieve accurate segmentation of fine vessel structures. Unlike CIDN, which focuses on attention mechanisms, CASR-Net redefines the decoder stage itself by replacing conventional convolutional layers with Self-ONN operators, enabling improved vessel continuity while maintaining computational efficiency. Complementary preprocessing with CLAHE and an enhanced Ben Graham method further improves performance on low-contrast angiograms. While prior studies such as Chang et al., Shen et al., and Li et al. report strong performance in terms of DSC, accuracy, or IoU, many rely on small test sets or omit cross-validation, limiting reproducibility. CASR-Net, by contrast, was validated using 5-fold cross-validation on a combined dataset containing both healthy and stenotic arteries. This ensures exposure to diverse vessel characteristics and supports better generalization. Given that stenosis is a major clinical concern and a central focus of coronary angiography, explicitly incorporating these cases in training and evaluation enables CASR-Net to deliver robust and clinically meaningful performance.

## 7. Limitations and Future Scopes

While CASR-Net demonstrates promising results in segmenting coronary arteries from angiograms, several limitations must be acknowledged to guide future research. One of the primary limitations is the availability and diversity of datasets. The current study relies on a limited number of annotated datasets, which may not fully reflect the variability in real-world clinical settings. This can affect the model's ability to generalize across different populations, imaging devices, and acquisition conditions. Moreover, since the evaluation in this study is limited to internal cross-validation, the model’s expected generalizability to external datasets and varying imaging conditions remains to be validated, which is critical for establishing clinical applicability. Additionally, the model is designed to process individual frames, without considering temporal information across angiographic sequences. This lack of temporal modelling restricts its application in dynamic or video-based diagnostic scenarios. Another limitation occurs during refinement, during which heuristic, rule-based methods are used to clean up segmentation outputs. Although these methods are generally effective, they may not be consistent across a variety of imaging conditions or anatomical variations. Furthermore, our current evaluation primarily relies on global metrics such as Dice and IoU, which do not fully capture vessel continuity in thin branches or stenotic regions. While we reported clDice for the top-performing models as an additional analysis, this evaluation was not extended across all networks. In addition, the use of manual annotations during training introduces potential variability and observer bias, which is not explicitly accounted for in the model. The preprocessing pipeline also follows a fixed strategy, which may not be optimal for all types of input images, particularly those with extreme lighting or noise. Another limitation is that the study did not include heavier architectures such as CiT-Net, which are computationally demanding but have shown promising results in related segmentation tasks.

To overcome these challenges, future studies should evaluate CASR-Net on larger, multi-center datasets that include a broader range of cases and imaging parameters. In particular, validating the model on external datasets with diverse imaging protocols would provide stronger evidence of robustness and clinical applicability across varied settings. Incorporating temporal modelling techniques, such as recurrent networks or transformer-based video models, could enhance performance on sequential data. Future work should also systematically incorporate topology-aware metrics such as clDice and Skeleton Recall Loss across all networks to provide a more complete assessment of vessel continuity. Future work will also explore the integration of heavier transformer-based models, such as CiT-Net, to investigate whether their greater representational capacity can further improve vessel continuity and robustness. Replacing the current heuristic refinement stage with a learnable, adaptive module may improve generalization. Semi-supervised or self-supervised learning strategies could help reduce the reliance on manually labelled data, while uncertainty estimation methods could increase model transparency and clinical trust. Together, these improvements can make CASR-Net more robust, adaptable, and clinically viable.

## 8. Conclusion

This study addressed key challenges in coronary artery segmentation by introducing CASR-Net, a novel pipeline integrating innovations in preprocessing, segmentation, and refinement. The multichannel preprocessing strategy, which combines CLAHE with an improved Ben Graham method, enhanced vessel visibility in noisy and low-contrast angiograms. The segmentation architecture extended the UNet framework with a Self-ONN–based decoder, allowing dynamic neuron adaptation during training and improving continuity in thin and stenotic vessels. The refinement stage further improved mask quality by suppressing false positives and reconnecting fragmented branches through patch-line generation. CASR-Net was rigorously validated using 5-fold cross-validation and achieved an IoU of 61.43%, a DSC of 76.10%, and a clDice of 79.36%, outperforming several state-of-the-art baselines. These results demonstrate that CASR-Net offers a robust and effective solution for coronary artery segmentation and holds strong potential to support clinical decision-making in the diagnosis and management of coronary artery disease.

## Supplementary Materials

- **Supplementary Table 1:** CLAHE Parameter Selection. This table details the quantitative results (Dice scores) from the ablation study, evaluating various combinations of CLAHE clip limits (4.0, 8.0, and 12.0) and grid sizes (4x4, 8x8, and 12x12).

## Declarations:

**Conflicts of interest:** The authors declare no conflict of interest.

**Data Availability Statement:** This work has utilized two open-source datasets such as Database X-ray Coronary Angiograms (DCA1)(https://doi.org/10.3390/app9245507) and Angiographic dataset for stenosis detection (https://data.mendeley.com/datasets/ydrm75xywg/1).

**Funding:** This work was made possible by High Impact grant of Qatar University # QUHI-CENG-22_23-548 and is also supported via funding from Prince Sattam Bin Abdulaziz University project number (PSAU/2024/R/1445). The statements made herein are solely the responsibility of the authors. The open-access publication cost is covered by Manipal Academy of Higher Education, United Arab Emirates.

**Ethics Statement:** Primary data collection for this study was conducted in accordance with the Declaration of Helsinki, ensuring the rights, safety, and confidentiality of participants. The present study, however, only utilized secondary data and had no direct interaction with humans.

**Clinical trial number:** Not applicable

**Human Ethics and Consent to Participate declarations:** Not applicable

**Acknowledgment:** A preprint of this paper has previously been published (Hassan et al., 2025).

**Author Contribution:** A.H & R.S– Methodology, Software, Writing – Original draft; M.E.C.H – Conceptualization, Methodology, Resources, Funding acquisition; M.M – Conceptualization, Methodology, Investigation, Validation, Review & Rewrite – Original draft; N.R.S – Investigation, Validation, Funding acquisition; A.A - Data curation, Funding acquisition, Writing - Original draft; S.B.Z – Visualization, Validation; B.B – Methodology, Validation, Supervision